\documentclass[journal]{IEEEtran}

\usepackage{amsmath,amsfonts,amssymb,mathtools,amsthm}
\usepackage{algorithmic}
\usepackage{array}
\usepackage{textcomp}
\usepackage{stfloats}
\usepackage{url}
\usepackage{verbatim}
\usepackage{graphicx}
\usepackage{balance}
\usepackage[hypertexnames=false]{hyperref}
\usepackage[numbers,sort&compress]{natbib}
\usepackage{booktabs}
\usepackage{microtype}
\usepackage{xcolor}
\usepackage{makecell}
\usepackage{multirow}
\usepackage{xspace}
\usepackage{tabularx}
\usepackage{wrapfig}
\usepackage{float}
\usepackage{color,colortbl}
\usepackage{pifont}
\usepackage[absolute,overlay]{textpos}

\theoremstyle{plain}

\theoremstyle{definition}

\theoremstyle{remark}

\newcommand{\cmark}{\ding{51}}
\newcolumntype{Y}{>{\centering\arraybackslash}X}

\def\ie{\emph{i.e., }}
\def\eg{\emph{e.g., }}
\newcommand{\ourmethod}{{\textup{\textsc{Pref-GRPO}}}\xspace}
\newcommand{\ourbench}{{\textup{\textsc{UniGenBench}}}\xspace}

\def\BibTeX{{\rm B\kern-.05em{\sc i\kern-.025em b}\kern-.08em
    T\kern-.1667em\lower.7ex\hbox{E}\kern-.125emX}}

\begin{document}

\title{Pref-GRPO: Pairwise Preference Reward-based GRPO for Stable Text-to-Image Reinforcement Learning}

\author{\textbf{Yibin Wang}\textsuperscript{1,2,3*}, \textbf{Zhimin Li}\textsuperscript{3*}, \textbf{Yuhang Zang}\textsuperscript{4*}, \textbf{Yujie Zhou}\textsuperscript{4,5}, \textbf{Jiazi Bu}\textsuperscript{4,5}  \\
\textbf{Chunyu Wang}\textsuperscript{3\dag}, \textbf{Qinglin Lu}\textsuperscript{3\dag}, \textbf{Cheng Jin}\textsuperscript{1,2\dag}, \textbf{Jiaqi Wang}\textsuperscript{2\dag}

\textsuperscript{1}Fudan University, \textsuperscript{2}Shanghai Innovation Institute
\textsuperscript{3}Hunyuan, Tencent, \\ \textsuperscript{4}Shanghai AI Lab, \textsuperscript{5}Shanghai Jiao Tong University \\

\textbf{Project Page}: \href{https://codegoat24.github.io/UnifiedReward/Pref-GRPO}{codegoat24.github.io/UnifiedReward/Pref-GRPO}

\thanks{\textsuperscript{*}{Equal contribution. \textsuperscript{\dag}Corresponding author.}}}

\markboth{JOURNAL OF \LaTeX\ CLASS FILES}%
{Pref-GRPO: Pairwise Preference Reward-based GRPO}

\twocolumn[{%
\renewcommand\twocolumn[1][]{#1}%
\maketitle
\vspace{-1em}

  \refstepcounter{figure}
  \includegraphics[width=1\textwidth]{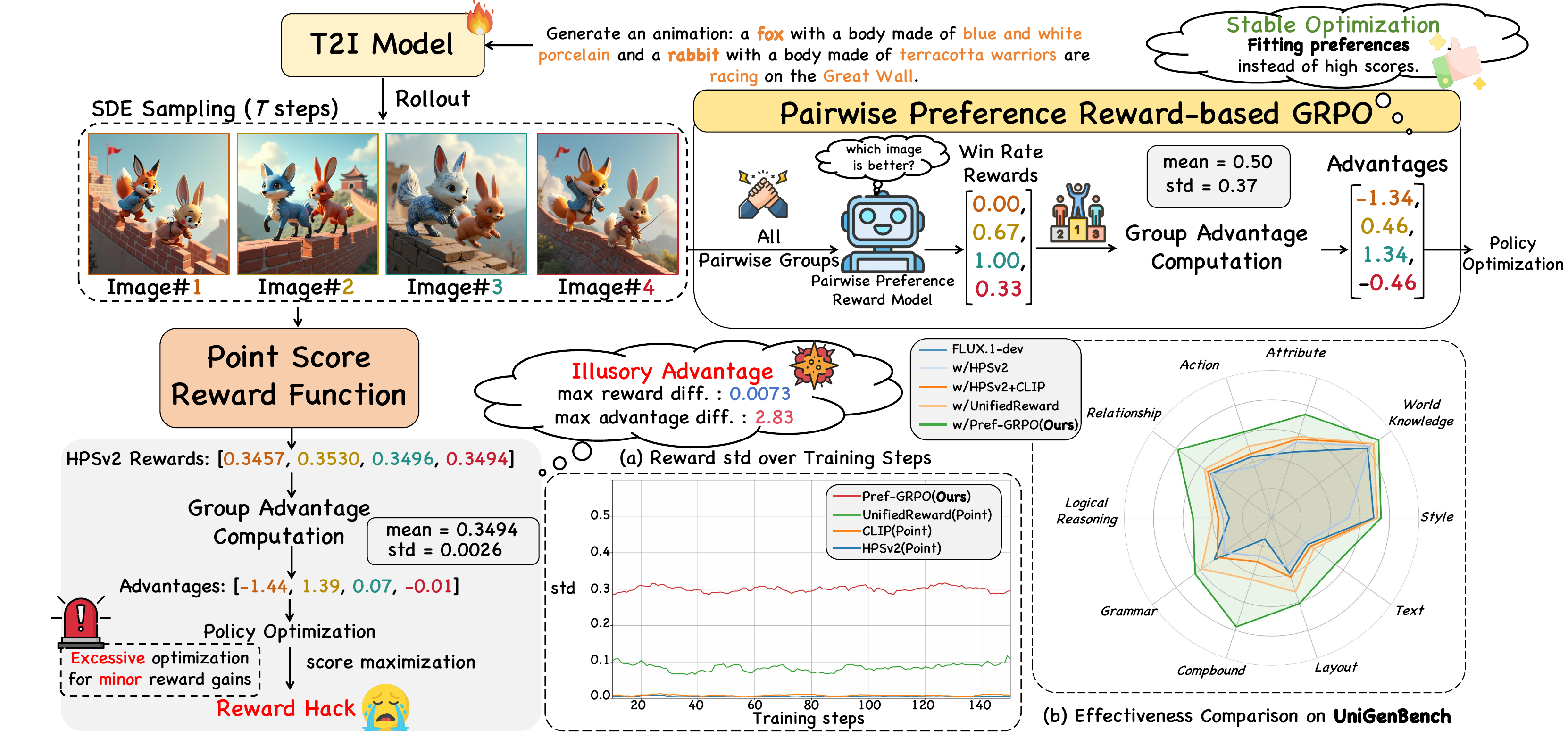}
  
  \label{fig:teaser}
\vspace{0.2em}

  \small \textbf{Fig. \thefigure. \ourmethod: from illusory advantage to pairwise preference fitting.} (a) Pointwise reward models assign tightly clustered scores to images within a group, producing an \textit{illusory advantage} after normalization and causing reward hacking. (b) \ourmethod reformulates the GRPO objective from absolute score maximization to pairwise preference fitting, enabling stable T2I optimization.
}]

\begin{abstract}
Recent progress has made GRPO-based reinforcement learning central to advancing text-to-image (T2I) generation. Current GRPO methods score a group of generated images with pointwise reward models and normalize the scores within each group to compute advantages. Although effective in early training, this reward-score-maximization approach is susceptible to \textbf{reward hacking}: scores keep rising while image quality deteriorates, manifesting as oversaturation or unnaturally dark artifacts. We trace this failure to an \textit{illusory advantage}: tightly clustered scores within a group are disproportionately amplified after dividing by the small group standard deviation, driving the policy to over-optimize. To address this, we propose \textbf{\ourmethod}, which reformulates the GRPO objective to pairwise preference fitting: image pairs within a group are compared by a Pairwise Preference Reward Model (PPRM), and each image's win rate serves as the reward. Because win rates reflect relative rankings rather than absolute scalar scores, \ourmethod yields larger within-group variance and is intrinsically robust to small biases. Experiments show that \ourmethod produces more stable advantages than pointwise scoring, substantially alleviates reward hacking, and improves semantic alignment.

Alongside the optimization side, existing T2I benchmarks offer only \textbf{coarse, primary-dimension-only evaluation}, reporting aggregate scores that hide where models actually succeed or fail. We propose \textbf{\ourbench}, a fine-grained T2I evaluation benchmark with 600 prompts across 5 primary themes and 20 sub-themes, evaluating 10 primary dimensions and 27 sub-dimensions with 1--5 explicit testpoints per prompt. Each testpoint is paired with a structured description, enabling precise per-testpoint assessment. Leveraging an MLLM (Gemini-2.5-Pro), we build an automated pipeline for benchmark construction and evaluation. Benchmarking representative open- and closed-source T2I models on \ourbench, we find that \emph{Style} and \emph{World Knowledge} are largely saturated across strong models, while fine-grained compositional capabilities such as \emph{Logical Reasoning}, \emph{Grammar}, and \emph{Compound} remain the primary bottlenecks---a diagnostic gap that aggregate scores cannot surface.

Together, \ourmethod and \ourbench contribute a more stable optimization objective and a more diagnostic evaluation framework, pointing to fine-grained compositional reasoning as the next frontier for T2I reinforcement learning.

\end{abstract}


\section{Introduction}
\IEEEPARstart{R}{ecent} progress in text-to-image (T2I) generation has made reinforcement learning \cite{flowgrpo,li2025mixgrpo,xue2025dancegrpo,he2025tempflow} and comprehensive benchmarking \cite{ghosh2023geneval,t2i-compbench,wei2025tiif} two pivotal pillars for advancing model capabilities and for evaluating them reliably.
Specifically, several GRPO-based approaches \cite{flowgrpo,xue2025dancegrpo} employ pointwise reward models \cite{unifiedreward,hpsv2,pickscore} to score a group of generated images at each training step and normalize these scores into advantages following the GRPO objective \cite{guo2025deepseek}. This pipeline has proven highly effective in aligning T2I generation with human preferences.

In parallel, evaluating T2I models, particularly their instruction-following capability, has become increasingly important. Widely adopted benchmarks \cite{t2i-compbench,ghosh2023geneval} probe compositional aspects using CLIP \cite{clip}-based or detection-based metrics, and TIIF-Bench \cite{wei2025tiif} further incorporates additional dimensions such as text rendering.

\begin{figure}[t]
    \centering
    \includegraphics[width=1\linewidth]{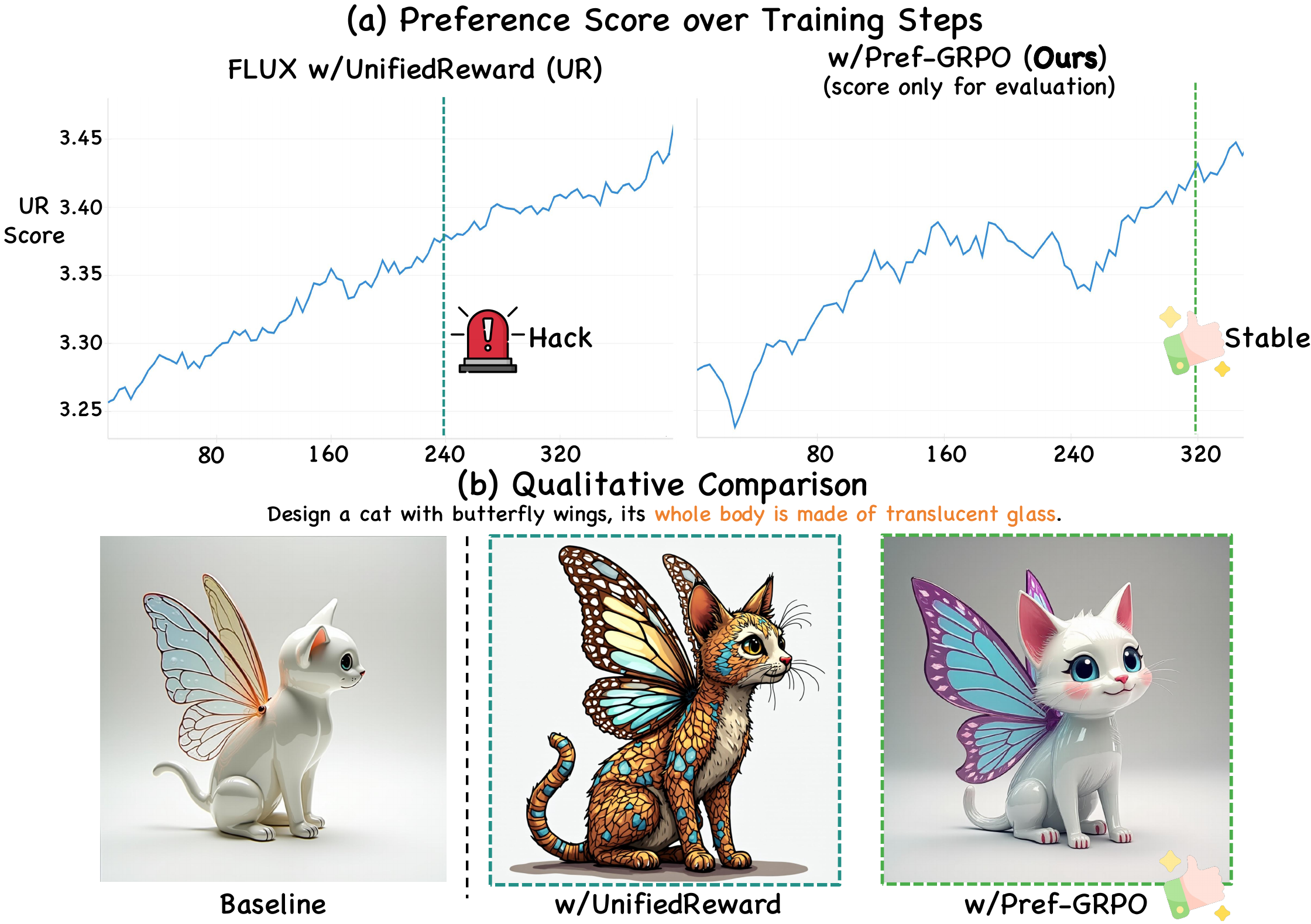}
    \vspace{-6pt}
    \caption{\textbf{Reward hacking under UnifiedReward.} (a) UnifiedReward scores rise monotonically during training. (b) Generated images gradually drift toward an unnaturally dark style, despite the rising reward, which is a clear reward-hacking signature.}
    \label{fig:reward_hack}
\end{figure}

Despite these advances, existing approaches still face two key limitations:
\textbf{(1)} Existing GRPO-based methods use pointwise reward models to maximize reward scores, which provides early gains but often leads to \textbf{reward hacking}: reward scores keep rising while image quality deteriorates over prolonged training (Fig.~\ref{fig:reward_hack}), a failure mode also noted in prior work \cite{flowgrpo,xue2025dancegrpo}.
\textbf{(2)} Current T2I benchmarks provide only \textbf{coarse, primary-dimension-only evaluation}, covering a limited range of sub-dimensions without reporting fine-grained sub-dimension scores (Tab.~\ref{tab:bench_compare}).

In light of these issues, we argue that \textbf{(1)} reward hacking in GRPO-based methods stems from an \textit{illusory advantage} that arises when pointwise reward models assign tightly clustered scores to images within the same group. Normalizing these scores into advantages disproportionately amplifies small gaps. Under a reward-maximization objective,
such inflated advantages drive the policy to over-optimize for trivial reward cues, eventually steering it toward reward-hacking behaviors that rapidly increase scores while destabilizing generation (Figs.~\ref{fig:teaser} and \ref{fig:reward_hack}). Moreover, small biases in the reward model are similarly amplified, encouraging the policy to exploit reward-model flaws rather than align with human preferences.
\textbf{(2)} Current T2I models already perform well on most primary dimensions (\eg object attributes and actions), which makes it necessary to decompose these broad dimensions into finer-grained sub-dimensions for more rigorous, diagnostic evaluation.

To this end, this work proposes \ourmethod, the first pairwise-preference-reward-based GRPO method for stable T2I reinforcement learning, and \ourbench, a fine-grained T2I evaluation benchmark for semantic consistency.

\textbf{(1)} \textbf{\ourmethod} incorporates a Pairwise Preference Reward Model (PPRM) \cite{unifiedreward-think}, reformulating the GRPO objective from absolute reward score maximization to pairwise preference fitting. As illustrated in Fig.~\ref{fig:teaser}, at each training step, given a group of generated images, we use the PPRM on each image pair to identify the preferred image. Each image's \textit{win rate} (the proportion of pairwise comparisons in which it is preferred) is then used as the reward signal for policy optimization. This design offers three key advantages:
(a) \textbf{Amplified reward variance}: driving high-quality images toward win rates near 1 and low-quality ones toward 0 yields more separable reward distributions and more informative advantage estimates.
(b) \textbf{Robustness to reward noise}: relying on relative rankings rather than absolute scores reduces over-optimization for marginal score gains and mitigates reward hacking.
(c) \textbf{Alignment with human preference}: pairwise comparisons mirror the way humans compare images, producing reward signals that better capture nuanced preferences. Extensive experiments show that \ourmethod discerns subtle quality variations, produces more stable and directional advantages than pointwise scoring, and alleviates reward hacking.
\begin{figure}[t]

    \centering
    \includegraphics[width=1\linewidth]{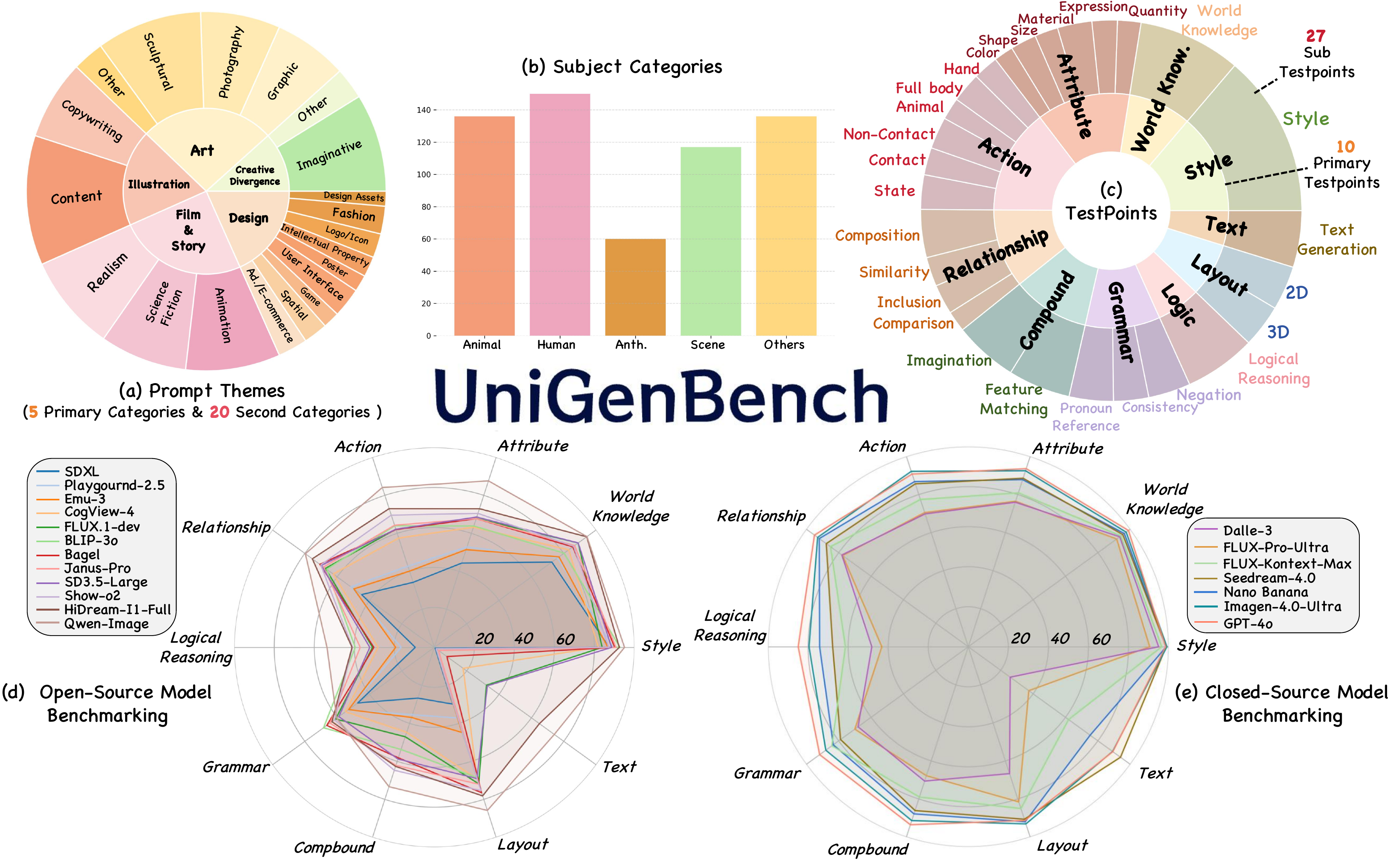}
    \vspace{-6pt}
    \caption{\textbf{\ourbench overview.} (a) Prompt themes: 5 primary categories and 20 sub-categories. (b) Subject categories. (c) Evaluation testpoints: 10 primary dimensions and 27 sub-dimensions. (d,e) Benchmarking results of representative open-source and closed-source T2I models on \ourbench.}
    \label{fig:dataset}

\end{figure}

\begin{table*}[t]
    \caption{\textbf{Benchmark comparison with existing T2I benchmarks.} We summarize benchmark coverage across the dimensions in \ourbench. Existing benchmarks mainly report primary-dimension scores or cover only a sparse subset of sub-dimensions, whereas \ourbench provides 10 primary dimensions and 27 sub-dimensions with fine-grained per-sub-dimension evaluation.}
    \label{tab:bench_compare}
    \vspace{-6pt}
    \centering
    \scriptsize
    \setlength{\tabcolsep}{2.2pt}
    \renewcommand{\arraystretch}{1.08}
    \resizebox{\textwidth}{!}{%
    \begin{tabular}{cccccccccccccccccccccccccccccc}
    \toprule
    \textbf{Benchmark} & \makecell{\textbf{Score}\\\textbf{Mode}}
    & \textbf{Style} & \makecell{\textbf{World}\\\textbf{Know.}}
    & \multicolumn{6}{c}{\textbf{Attribute}}
    & \multicolumn{6}{c}{\textbf{Action}}
    & \multicolumn{4}{c}{\textbf{Relationship}}
    & \multicolumn{2}{c}{\textbf{Compound}}
    & \multicolumn{3}{c}{\textbf{Grammar}}
    & \multicolumn{2}{c}{\textbf{Layout}}
    & \makecell{\textbf{Logical}\\\textbf{Reason.}}
    & \textbf{Text} \\
    \cmidrule(lr){3-4}
    \cmidrule(lr){5-10}
    \cmidrule(lr){11-16}
    \cmidrule(lr){17-20}
    \cmidrule(lr){21-22}
    \cmidrule(lr){23-25}
    \cmidrule(lr){26-27}
    & &
    --- & ---
    & \textbf{Quant.} & \textbf{Expn.} & \textbf{Material} & \textbf{Size} & \textbf{Shape} & \textbf{Color}
    & \textbf{Hand} & \makecell{\textbf{Full}\\\textbf{Body}} & \textbf{Animal} & \makecell{\textbf{Non}\\\textbf{Contact}} & \textbf{Contact} & \textbf{State}
    & \textbf{Compo.} & \textbf{Similarity} & \textbf{Inclusion} & \textbf{Comparison}
    & \textbf{Imagin.} & \makecell{\textbf{Feature}\\\textbf{Matching}}
    & \makecell{\textbf{Pronoun}\\\textbf{Ref.}} & \textbf{Consistency} & \textbf{Negation}
    & \textbf{2D} & \textbf{3D}
    & --- & --- \\
    \midrule
    \makecell{GenEval} &  &  &  & \cmark &  &  &  &  & \cmark &  &  &  &  &  &  &  &  &  &  &  & \cmark &  &  &  & \cmark &  &  &  \\
    \makecell{T2I-Comp} & \makecell{Primary\\Dimension} &  &  & \cmark &  & \cmark &  & \cmark & \cmark &  & \cmark & \cmark & \cmark & \cmark & \cmark &  &  &  &  &  & \cmark &  &  &  & \cmark & \cmark &  &  \\
    \makecell{TIIF-Bench} & & \cmark & \cmark & \cmark &  & \cmark & \cmark & \cmark & \cmark &  & \cmark & \cmark & \cmark & \cmark & \cmark &  &  &  & \cmark & \cmark & \cmark &  &  & \cmark & \cmark & \cmark &  & \cmark \\
    \midrule
    \makecell{\textbf{UniGenBench}\\\textbf{(Ours)}} & \makecell{Primary\\\& Sub\\Dimension} & \cmark & \cmark & \cmark & \cmark & \cmark & \cmark & \cmark & \cmark & \cmark & \cmark & \cmark & \cmark & \cmark & \cmark & \cmark & \cmark & \cmark & \cmark & \cmark & \cmark & \cmark & \cmark & \cmark & \cmark & \cmark & \cmark & \cmark \\
    \bottomrule
    \end{tabular}%
    }
\vspace{-0.4cm}
\end{table*}

\textbf{(2)} \textbf{\ourbench} is built for fine-grained T2I evaluation, covering 10 primary dimensions and 27 sub-dimensions, alongside 5 prompt themes (20 sub-themes) and diverse subject categories (Fig.~\ref{fig:dataset}). Unlike existing benchmarks that provide only coarse, primary-dimension-only evaluation, most primary dimensions in \ourbench are further subdivided into fine-grained sub-dimensions, each associated with explicit testpoints (Tab.~\ref{tab:bench_compare}). We further build an automated evaluation pipeline powered by a strong MLLM (Gemini-2.5-Pro~\cite{huang2025gemini}) for both benchmark construction and T2I model evaluation, illustrated in Fig.~\ref{fig:bench_pipeline}.
We benchmark popular closed-source models, including GPT-4o \cite{hurst2024gpt}, Nano Banana, Seedream-4.0 \cite{gao2025seedream}, and Imagen-4.0-Ultra \cite{imagen}, as well as leading open-source models such as Qwen-Image \cite{qwen_image}, HiDream-I1-Full \cite{cai2025hidream}, and Bagel \cite{deng2025emerging}. As shown in Fig.~\ref{fig:dataset}(d, e), both open- and closed-source models perform relatively well on \emph{Style} and \emph{World Knowledge} prompts, but consistently underperform on prompts requiring \emph{Logical Reasoning}, such as those involving causal or contrastive relationships.

\textbf{Contributions}:
\textbf{(1)} We identify a root cause of reward hacking in pointwise-reward GRPO: the \textit{illusory advantage} induced by within-group score normalization.
\textbf{(2)} We propose \ourmethod, the first pairwise-preference-reward-based GRPO method for stable T2I reinforcement learning, which reformulates the optimization objective from absolute reward score maximization to pairwise preference fitting.
\textbf{(3)} Experiments on \ourbench, GenEval, and T2I-CompBench show that \ourmethod alleviates reward hacking and improves semantic alignment (+5.84\% overall on \ourbench, with +12.69\% on \textit{Text} and +12.04\% on \textit{Logical Reasoning}), without sacrificing perceptual quality.
\textbf{(4)} We introduce \ourbench, a fine-grained T2I benchmark with 10 primary dimensions and 27 sub-dimensions across 5 prompt themes (20 sub-themes), together with an MLLM-based pipeline for benchmark construction and evaluation.
\textbf{(5)} We benchmark representative open- and closed-source T2I models on \ourbench, revealing that \emph{Style} and \emph{World Knowledge} are saturated across strong models, while fine-grained compositional capabilities such as \emph{Logical Reasoning}, \emph{Grammar}, and \emph{Compound} remain the primary bottlenecks.

\section{Related Work}

\textbf{Reinforcement Learning for T2I Generation} is gaining rapid momentum. Early efforts pursued preference-driven objectives \cite{xie2025dymo,yang2024using,wallace2024diffusion}. 
For example, DiffusionDPO \cite{wallace2024diffusion}, adapted from Direct Preference Optimization (DPO) \cite{rafailov2023direct}, optimizes preference consistency via a classification-style objective as a simpler alternative to RLHF. 
DyMO \cite{xie2025dymo} offers a plug-and-play, training-free alignment method that steers generated images toward human preferences at inference. 
More recently, group relative policy optimization (GRPO) has advanced online RL-enhanced image generation. Flow-GRPO \cite{tong2025delving} and DanceGRPO \cite{xue2025dancegrpo} instantiate GRPO on flow-matching models, introducing stochasticity by recasting the original deterministic ODE as an equivalent SDE. 
While effective, these reward score-maximization methods are prone to reward hacking due to illusory advantage. To this end, we propose \ourmethod, which shifts training from reward-score maximization to pairwise preference fitting, yielding more stable advantages, and thereby mitigates reward hacking.

\textbf{Existing Benchmarks for T2I Evaluation} have expanded the evaluation of T2I models beyond simple visual fidelity, incorporating compositional reasoning \cite{ghosh2023geneval,t2i-compbench} and world knowledge~\cite{niu2025wise}. 
Recently, \cite{wei2025tiif} introduces TIIF-Bench, containing 5k prompts spanning multiple dimensions, \ie text rendering and style control, rigorously evaluating model robustness to variations in prompt length.
However, existing benchmarks largely focus on primary dimension–level coarse assessment, covering a limited set of sub-tasks and lacking fine-grained
assessment of these sub-tasks. To address this gap, we propose a unified image generation benchmark, \ourbench, consisting of 600 prompts spanning diverse themes and categories, assessing T2I models across 10 primary dimensions and 27 sub-dimensions.

\section{Motivation for Pairwise Preference Reward}

Existing GRPO-based text-to-image methods optimize the policy by maximizing \emph{pointwise} reward scores \cite{flowgrpo,xue2025dancegrpo,li2025mixgrpo,he2025tempflow}. This design implicitly assumes that the absolute score assigned by the reward model is both sufficiently discriminative within a rollout group and reliable enough to support gradient scaling. In practice, however, these assumptions are often violated. For visually similar images sampled from the same prompt, pointwise reward models such as HPSv2 \cite{hpsv2} and UnifiedReward \cite{unifiedreward} frequently produce highly compressed scores (see Fig. \ref{fig:teaser}). GRPO then normalizes these near-identical scores by a very small group standard deviation, turning negligible score gaps into disproportionately large advantages. The optimization is therefore driven not by substantial quality differences, but by amplified numerical fluctuations, which leads to the \emph{illusory advantage} phenomenon, as analyzed in Sec. \ref{sec:adv_exp}.

The consequence is not merely optimization instability, but a deeper objective mismatch. Once the training target is defined as absolute reward maximization, the policy is encouraged to chase any reward increase, even when that increase comes from reward-model bias rather than genuine improvement in text-image alignment or visual quality. This helps explain the reward hacking behaviors reported in prior work \cite{flowgrpo,xue2025dancegrpo} and observed in our experiments: HPSv2 encourages over-saturated generations (see Fig. \ref{fig:hps_hacking}), while UnifiedReward tends to favor an unnatural dark style (see Fig. \ref{fig:reward_hack}). Importantly, simply shrinking or clipping the normalized advantage can only dampen the update magnitude; it cannot remove the underlying tendency to optimize toward unreliable absolute-score increments.

These observations motivate replacing reward \emph{score maximization} with \emph{pairwise preference fitting}. Relative comparison is fundamentally better suited to this setting for two reasons. First, when candidate images are close in quality, determining which one is better is typically more reliable than assigning well-calibrated absolute scores to each image independently. Second, pairwise comparison discards uninformative score magnitude and focuses optimization on stable ordering information, making the reward signal less sensitive to small fluctuations and common calibration bias. This change also better matches human evaluation, which is usually comparative rather than absolute for similar generations. Motivated by this, we reformulate traditional GRPO with pairwise preference rewards and pairwise win rates, as detailed next.

\section{\ourmethod}
This section presents the technical formulation of \ourmethod. We first introduce GRPO on flow matching models in Sec.~\ref{sec:flow_matching}, then analyze the \textit{illusory advantage} phenomenon in Sec.~\ref{sec:adv_exp}, and finally describe our \ourmethod objective in Sec.~\ref{sec:pref_grpo}.

\subsection{Flow Matching GRPO} \label{sec:flow_matching}

\paragraph{Flow Matching} 
Let $x_0 \sim p_{\text{data}}$ be a data sample and $x_1 \sim \mathcal{N}(0,I)$ a noise sample. Rectified flow \cite{liu2022flow} defines intermediate samples as
\begin{equation}
x_t = (1-t)x_0 + t x_1, \quad t \in [0,1],
\end{equation}
and trains a velocity field $v_\theta(x_t,t)$ via the flow matching \cite{lipman2022flow} objective:
\begin{equation}
\mathcal{L}_{\text{FM}}(\theta) = \mathbb{E}_{t,x_0,x_1} \big[\|v - v_\theta(x_t,t)\|_2^2 \big], \quad v = x_1 - x_0.
\end{equation}

Beyond training, the iterative denoising process at inference time can be naturally formalized as a Markov Decision Process \cite{black2023training}. We discretize $t\in[0,1]$ into $T$ steps. At each step, the state is $s_t = (c, t, x_t)$, where $c$ denotes the prompt. The action $a_t$ corresponds to producing the denoised sample $x_{t-1} \sim \pi_\theta(x_{t-1}|x_t,c)$, yielding a deterministic transition to $s_{t+1} = (c, t-1, x_{t-1})$. The initial state samples a prompt $c \sim p(c)$ with $t=T$ and $x_T \sim \mathcal{N}(0,I)$. A reward is only provided at the terminal state ($t{=}0$): $R(x_0,c)$, and zero otherwise.

\paragraph{GRPO on Flow Matching} 
GRPO \cite{guo2025deepseek} introduces a group-relative advantage to stabilize policy updates. On flow matching models, for a group of $G$ generated images $\{x_0^i\}_{i=1}^G$, the advantage of the $i$-th image is
\begin{equation}
\hat{A}_t^i = \frac{R(x_0^i,c) - \mathrm{mean}(\{R(x_0^j,c)\}_{j=1}^G)}{\mathrm{std}(\{R(x_0^j,c)\}_{j=1}^G)}. \label{eqa:adv}
\end{equation}
The policy is updated by maximizing the clipped, KL-regularized objective
\begin{equation}
\mathcal{J}_{\text{Flow}}(\theta) = \mathbb{E}_{c, \{x^i\}} \Big[ f(r, \hat{A}, \theta, \eta, \beta) \Big],
\end{equation}
where
\begin{equation}
\begin{aligned}
f(r, \hat{A}, \theta, \eta, \beta)
= \frac{1}{G}\sum_{i=1}^G \frac{1}{T} \sum_{t=0}^{T-1} 
\min\Big(
r_t^i(\theta)\hat{A}_t^i,\,\\
\text{clip}\big(r_t^i(\theta),1-\eta,1+\eta\big)\hat{A}_t^i
\Big)  - \beta D_{\text{KL}}\!\left(\pi_\theta \,\|\, \pi_{\text{ref}}\right).
\end{aligned}
\end{equation}
with $r_t^i(\theta) = \frac{p_\theta(x_{t-1}^i|x_t^i,c)}{p_{\theta_{\text{old}}}(x_{t-1}^i|x_t^i,c)}$.

To enable the stochastic exploration required by GRPO, Flow-GRPO~\cite{flowgrpo} converts the deterministic flow ODE $dx_t = v_\theta(x_t,t)\,dt$ to an equivalent SDE:
\begin{equation}
dx_t = \big(v_\theta(x_t,t) + \frac{\sigma_t^2}{2t} (x_t + (1-t)v_\theta(x_t,t))\big) dt + \sigma_t dw_t,
\end{equation}
where $\{w_t\}$ is a standard Wiener process and $\sigma_t$ is a time-dependent noise scale.  
Euler-Maruyama discretization gives the update rule:
\begin{equation}
\begin{aligned}
x_{t+\Delta t} = x_t + \big(v_\theta(x_t,t) + \frac{\sigma_t^2}{2t}(x_t + (1-t)v_\theta(x_t,t))\big)\Delta t \\+ \sigma_t \sqrt{\Delta t} \epsilon, \quad \epsilon \sim \mathcal{N}(0,I).
\end{aligned} 
\end{equation}
where $\sigma_t = a\sqrt{t/(1-t)}$ and $a$ is a scalar hyperparameter.

\subsection{Illusory Advantage in Pointwise-Reward GRPO} \label{sec:adv_exp}
Existing flow-matching-based GRPO methods \cite{flowgrpo,xue2025dancegrpo,li2025mixgrpo,he2025tempflow} use pointwise reward models \cite{unifiedreward,clip,hpsv2} to score a group of generated images at each update step. The advantage of each image is then obtained by normalizing its reward within the group, as in Eq.~\ref{eqa:adv}. This normalization is intended to standardize updates across samples. In practice, however, pointwise reward models often assign tightly clustered scores $R(x_0^i, c)$ to visually similar images from the same prompt, yielding a small within-group standard deviation $\sigma_r$. As a consequence, even minor score gaps are disproportionately magnified after normalization (see Fig.~\ref{fig:teaser}(a)).

To make this effect explicit, let $\mu_r$ and $\sigma_r$ denote the mean and standard deviation of the group's rewards. Define the centered reward of sample $i$ as $\Delta r_i = (x_0^i, c) - \mu_r$. The normalized advantage then becomes:
\begin{equation}
\hat{A}_t^i = \frac{\Delta r_i}{\sigma_r}.
\end{equation}

For clarity, we drop the clip and KL terms and isolate the effect of reward normalization. The gradient of the GRPO objective with respect to $\theta$ is then:\begin{equation}
\nabla_\theta \mathcal{J}_{\text{Flow}}(\theta)
= \mathbb{E}_{c, \{x^i\}} 
\Bigg[
\frac{1}{G} \sum_{i=1}^G \frac{1}{T}\sum_{t=0}^{T-1} 
\nabla_\theta \big( r_t^i(\theta) \hat{A}_t^i \big)
\Bigg].
\end{equation}

Substituting $\hat{A}_t^i = \Delta r_i / \sigma_r$ and treating $\sigma_r$ as a per-group constant, we obtain:
\begin{equation}
\begin{aligned}
\nabla_\theta \mathcal{J}_{\text{Flow}}(\theta)
= \frac{1}{\sigma_r} \,
\mathbb{E}_{c, \{x^i\}} 
\Bigg[
\frac{1}{G} \sum_{i=1}^G \frac{1}{T}\sum_{t=0}^{T-1} 
\nabla_\theta \big( r_t^i(\theta) \Delta r_i \big)
\Bigg].
\end{aligned}
\end{equation}

The gradient norm therefore contains an explicit factor of $1/\sigma_r$:
\begin{equation}
\|\nabla_\theta \mathcal{J}_{\text{Flow}}\| \propto \frac{1}{\sigma_r}.
\end{equation}

Therefore, when a pointwise reward model assigns nearly identical scores to images within the same group (\ie $R(x_0^i, c) \approx R(x_0^j, c)$), the normalization becomes highly sensitive to small perturbations. Even when $\Delta r_i$ is small, dividing by a small $\sigma_r$ produces disproportionately large normalized advantages $\hat{A}_t^i$, making the update direction overly dependent on noisy reward differences.

This amplification causes the model to over-optimize for spurious reward differences, manifesting as \textit{reward hacking}. We term this failure mode the \textit{illusory advantage} phenomenon.
The \textit{illusory advantage} has two detrimental effects. (1) \textbf{Excessive optimization}: minimal score variations are exaggerated, pushing the policy toward extreme, reward-hacked behaviors (Figs.~\ref{fig:reward_hack} and \ref{fig:hps_hacking}). (2) \textbf{Sensitivity to reward noise}: the optimization becomes highly susceptible to biases in the reward model, encouraging the policy to exploit model flaws rather than pursue genuine quality improvements.

\subsection{Pairwise Preference Reward-based GRPO} \label{sec:pref_grpo}
To mitigate the \textit{illusory advantage}, we propose \textbf{\ourmethod}, which uses a Pairwise Preference Reward Model (PPRM) \cite{unifiedreward-think} to reformulate the optimization objective as pairwise preference fitting. Instead of relying on absolute reward scores, \ourmethod evaluates relative preferences among generated images, mirroring the way humans compare images pairwise. This yields a reward signal that better captures nuanced differences in image quality and prompt alignment, producing more stable and informative advantages for policy optimization while reducing susceptibility to reward hacking.
Specifically, given a set of $G$ images $\{x_0^i\}_{i=1}^G$ sampled from $\pi_\theta(\cdot \mid c)$ for prompt $c$, for each ordered pair $(i,j)$ with $i \neq j$ we use the PPRM to determine whether image $i$ is preferred over image $j$. The \textit{win rate} of image $i$ is defined as
\begin{equation}
w_i = \frac{1}{G-1} \sum_{j \neq i} \mathbb{1} \big( x_0^i \succ x_0^j \big),
\end{equation}
where $\mathbb{1}(\cdot)$ is the indicator function, and $x_0^i \succ x_0^j$ indicates that image $i$ is preferred over image $j$ according to the PPRM. The win rates are then used as rewards, replacing the scalar rewards $R(x_0^i,c)$ in the GRPO advantage:
\begin{equation}
\hat{A}_t^i = \frac{w_i - \mathrm{mean}(\{w_j\}_{j=1}^G)}{\mathrm{std}(\{w_j\}_{j=1}^G)}.
\end{equation}
Compared to reward score maximization, \ourmethod offers several advantages:
(1) \textbf{Amplified reward variance}: replacing scalar rewards with pairwise win rates spreads win rates across $[0,1]$, yielding a larger within-group variance. High-quality samples approach $1$, and lower-quality samples approach $0$, producing a more discriminative and robust reward distribution for advantage estimation, thereby mitigating reward hacking.
(2) \textbf{Robustness to reward noise}: because optimization relies on relative rankings rather than raw scores, \ourmethod reduces sensitivity to small fluctuations and biases in the reward model, lowering the likelihood of reward hacking and improving training stability.
(3) \textbf{Alignment with human preference}: human preference judgments over images are inherently relative rather than absolute. Mirroring this process lets \ourmethod capture fine-grained quality distinctions often missed by pointwise scoring, providing a more faithful signal for policy improvement.

\section{\ourbench}
Rigorous evaluation is equally important for discerning whether a T2I model faithfully follows complex semantics. This requires a benchmark that goes beyond coarse aggregate scores and probes fine-grained semantic capabilities. However, existing benchmarks \cite{ghosh2023geneval,t2i-compbench,wei2025tiif} still exhibit the following limitations:
(1) \textbf{Sparse sub-dimension coverage}: existing benchmarks typically include only a few sub-dimensions within each primary dimension, missing many practical capabilities. For example, as shown in Tab.~\ref{tab:bench_compare}, current benchmarks include only a single sub-dimension for \textit{Relationship} and \textit{Grammar}, leading to incomplete and potentially misleading assessments in these aspects.
(2) \textbf{No sub-dimension-level reporting}: existing benchmarks report scores only at the primary-dimension level, without breaking down performance across sub-dimensions. This coarse reporting limits interpretability and prevents diagnosing a T2I model's strengths and weaknesses.

Motivated by this gap, we propose \textbf{\ourbench}, a T2I evaluation benchmark designed not only to compare models, but also to reveal where their semantic understanding actually breaks down. \ourbench offers the following advantages:
\begin{itemize}
\item \textbf{Comprehensive fine-grained evaluation}: 10 primary dimensions and 27 sub-dimensions, enabling substantially finer-grained diagnosis than benchmarks that report only coarse dimension-level scores.
\item \textbf{Broad prompt coverage}: 5 major prompt themes and 20 sub-themes, spanning scenarios from realistic rendering to open-ended creative composition.
\item \textbf{Efficient benchmark design}: 600 prompts, each with 1 to 5 explicit testpoints, achieving broad semantic coverage without the thousands of prompts used by prior benchmarks.
\item \textbf{Reliable MLLM-based assessment}: each prompt is paired with structured testpoint descriptions, enabling precise, interpretable per-testpoint assessment of whether the intended semantic targets are satisfied.
\end{itemize}
We will first introduce our design of prompt themes and evaluation criteria in the benchmark (Sec. \ref{sec:prompt_theme}), and then elaborate on our MLLM-based automated pipeline for prompt generation and T2I evaluation (Sec. \ref{sec:bench_pipeline}).
\begin{figure}[t]
    \centering
    \includegraphics[width=0.9
    \linewidth]{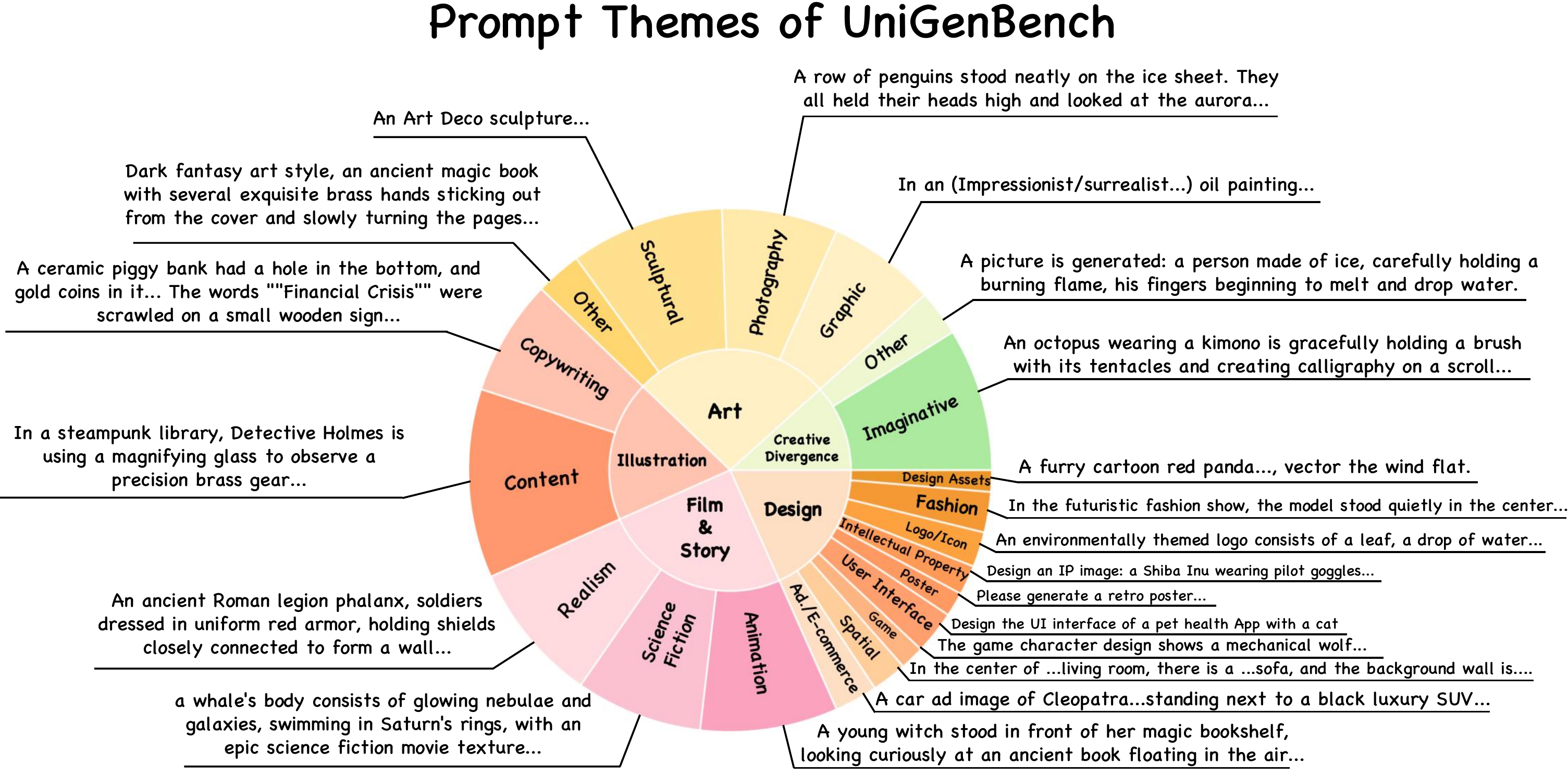}
    \caption{\textbf{Prompt themes of \ourbench.} Representative examples for each of the 5 primary themes and 20 sub-themes.}
    \label{fig:detailed_theme}
\end{figure}

\subsection{Prompt Themes and Evaluation Dimensions} \label{sec:prompt_theme}
As shown in Fig. \ref{fig:dataset}, \ourbench covers five major \textbf{prompt themes}: \textit{Art}, \textit{Illustration}, \textit{Creative Divergence}, \textit{Design}, and \textit{Film\&Storytelling}, further divided into 20 subcategories. It also spans diverse \textbf{subject categories} including \textit{animals}, \textit{objects}, \textit{anthropomorphic characters}, \textit{scenes}, and an \textit{Other} category for special entities (\eg robots in science-fiction themes). This design ensures the benchmark spans both practical generation scenarios and open-ended creative settings, rather than skewing toward a narrow prompt distribution.
Representative examples of the prompt themes are shown in Fig. \ref{fig:detailed_theme}.

\begin{figure}[t]

    \centering
    \includegraphics[width=1\linewidth]{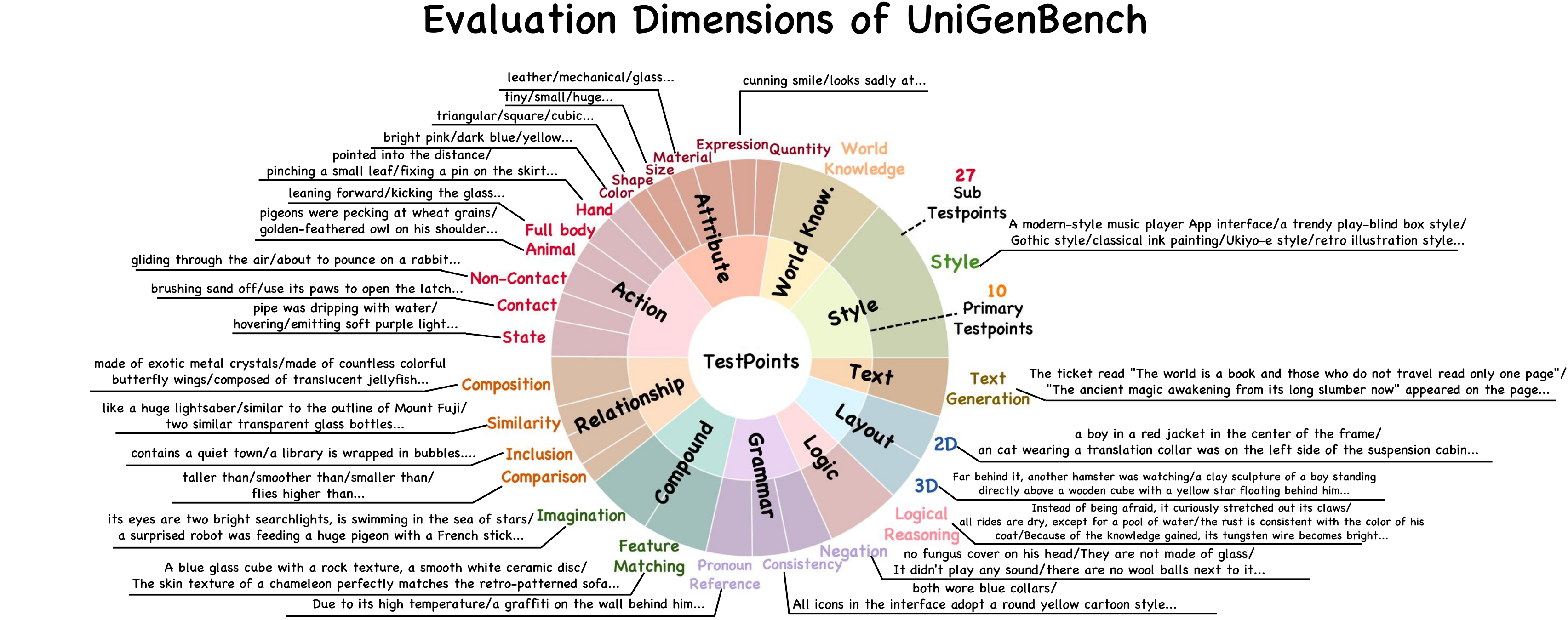}
    \caption{\textbf{Evaluation dimensions of \ourbench.} Representative prompt examples for each of the 27 sub-dimensions under the 10 primary evaluation dimensions.}
    \label{fig:detailed_testpoint}

\end{figure}

Unlike coarse benchmarks that only report a few high-level metrics, \ourbench defines 10 \textbf{primary evaluation dimensions} and 27 \textbf{sub-dimensions}. As illustrated in Fig. \ref{fig:detailed_testpoint}, we explicitly include several challenging yet practically important capabilities that are often overlooked, such as logical reasoning, facial expressions, pronoun reference, hand actions, and fine-grained relationship understanding (\eg composition, similarity, and inclusion relations), as well as grammatical consistency across multiple entities. This finer decomposition enables the benchmark to diagnose a model’s strengths and weaknesses more precisely, providing substantially richer feedback than dimension-level averages alone.

At a high level, these ten dimensions cover complementary aspects of semantic faithfulness: \textit{Style} evaluates adherence to the requested visual form; \textit{World Knowledge} evaluates consistency with commonsense and real-world facts; \textit{Attribute} evaluates whether entities exhibit the correct properties; \textit{Action} evaluates whether entities perform the intended behaviors; \textit{Relationship} evaluates whether inter-entity relations are grounded correctly; \textit{Layout} evaluates whether spatial arrangements are rendered accurately; \textit{Compound} evaluates whether multiple concepts are integrated coherently; \textit{Grammar} evaluates language-sensitive grounding such as reference, consistency, and negation; \textit{Logical Reasoning} evaluates causal, contrastive, and inferential understanding; and \textit{Text} evaluates whether explicit textual content is generated correctly.

To make these dimensions operational, we further instantiate them with explicit sub-dimension testpoints. Specifically, \textit{Attribute} is decomposed into quantity, expression, material, size, shape, and color; \textit{Action} is decomposed into hand actions, full-body motions, animal behaviors, contact interactions, non-contact interactions, and object states; \textit{Relationship} is decomposed into composition, similarity, inclusion, and comparison; and \textit{Layout} is decomposed into 2D placement and 3D spatial arrangement. This design allows \ourbench to reveal which aspect of semantic generation actually fails, rather than only reporting a coarse dimension-level average.

\begin{figure}[t]
    \centering
    \includegraphics[width=0.8\linewidth]{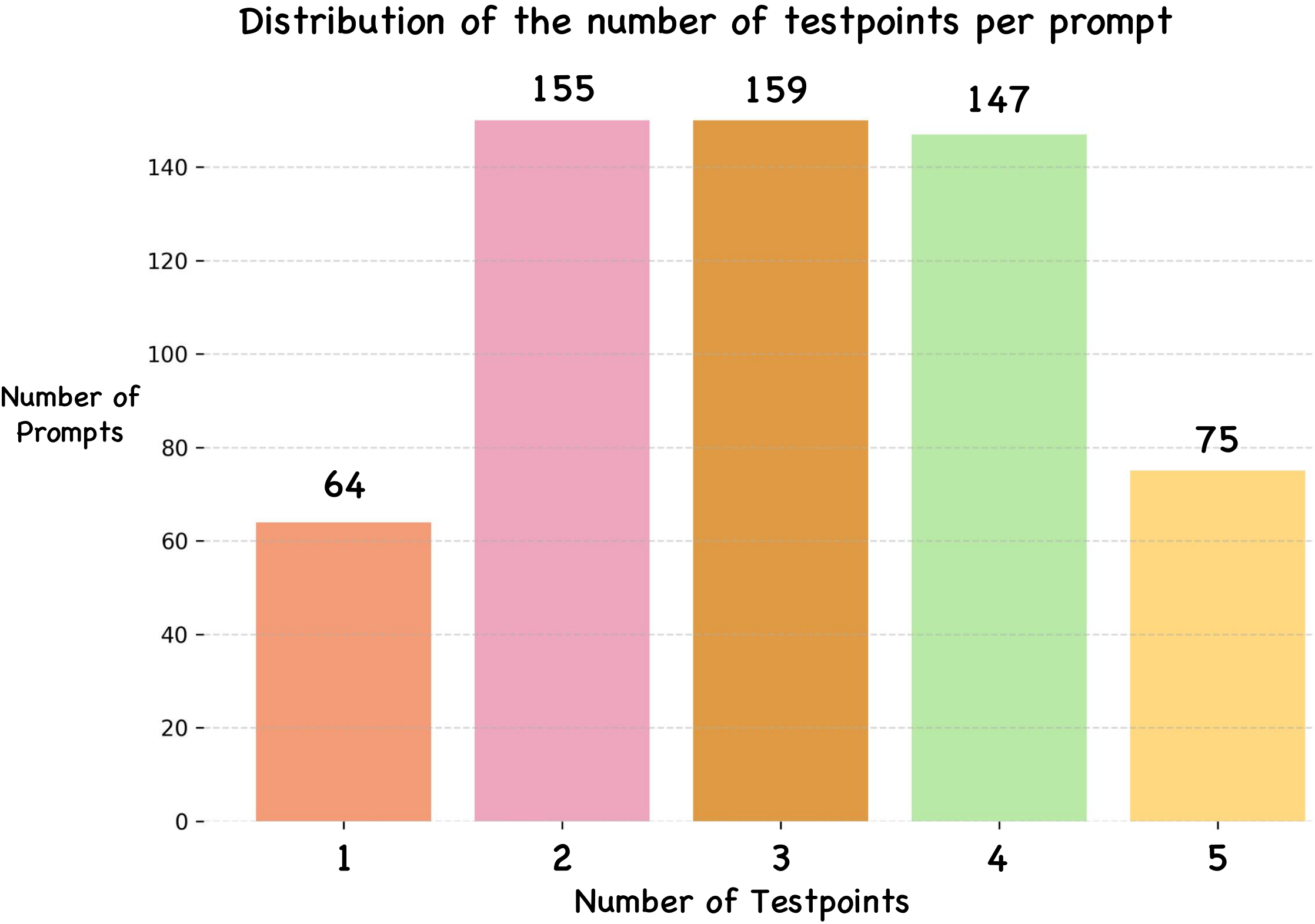}
    \caption{\textbf{Distribution of testpoint counts per prompt in \ourbench.} Each prompt is associated with 1 to 5 testpoints, balancing coverage and evaluation interpretability.}
    \label{fig:testpoint_num_distribution}
    \vspace{-0.3cm}
\end{figure}

We further control the complexity of each prompt through the number of associated testpoints. Instead of relying on thousands of prompts with diffuse evaluation targets, \ourbench uses 600 prompts, each focusing on 1 to 5 specific testpoints. As shown in Fig. \ref{fig:testpoint_num_distribution}, this design achieves a practical balance between coverage and efficiency: each prompt remains interpretable and evaluation-friendly, while the benchmark as a whole still spans a broad set of semantic capabilities.

\begin{figure}[t]

    \centering
    \includegraphics[width=1\linewidth]{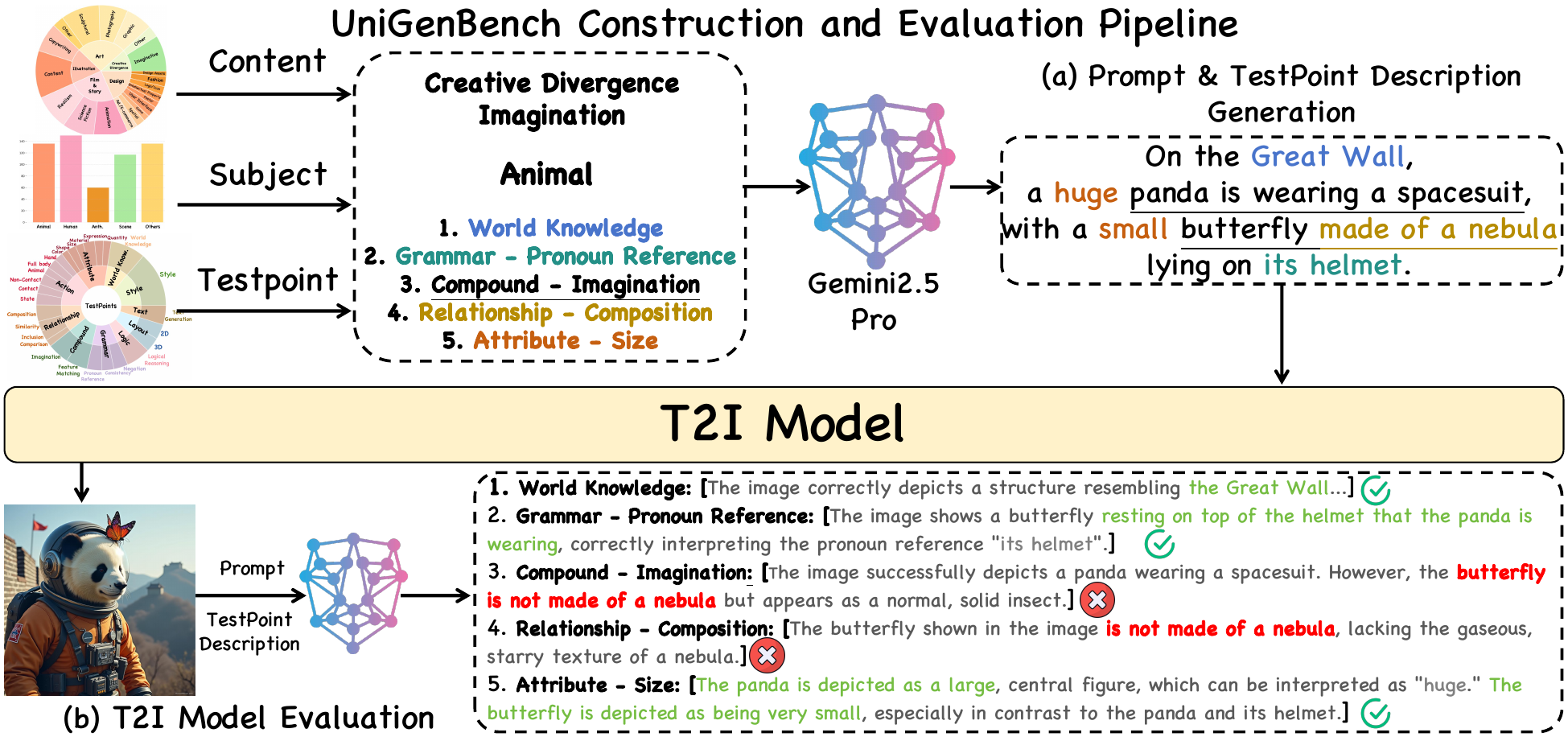}
    \caption{\textbf{\ourbench construction and evaluation pipeline.} An MLLM (Gemini-2.5-Pro) is used for (a) prompt and testpoint-description generation and (b) fine-grained per-testpoint evaluation.}
    \label{fig:bench_pipeline}
\end{figure}

\subsection{Benchmark Construction and Evaluation Pipeline} \label{sec:bench_pipeline}
Having established diverse prompt themes, subject categories, and evaluation dimensions, we further construct an MLLM-based automated pipeline to operationalize the benchmark shown in Fig. \ref{fig:bench_pipeline}. This pipeline serves two complementary purposes: (1) generating large-scale, diverse, and high-quality prompts in a systematic and controllable manner (Sec. \ref{sec:bench_construction}), and (2) enabling scalable, reliable, and fine-grained evaluation of T2I models (Sec. \ref{Sec:bench_eval}). By leveraging the reasoning and perception capabilities of MLLMs, the pipeline eliminates the need for costly human annotation, while ensuring both efficiency and reliability in benchmark construction and model assessment.

\subsubsection{Prompt and Testpoint Description Generation} \label{sec:bench_construction}
Let $\mathcal{T}$ denote the set of prompt \textit{themes}, $\mathcal{S}$ the set of \textit{subject categories}, and $\mathcal{C}$ the set of \textit{evaluation dimensions}. For each prompt, we sample a theme $t \sim \mathcal{T}$ and a subject category $s \sim \mathcal{S}$ uniformly at random. Subsequently, a subset of $k$ testpoints $\{c_1, \dots, c_k\} \subset \mathcal{C}$, with $k \in [1,5]$, is sampled to target specific fine-grained evaluation aspects.

The selected tuple $(t, s, \{c_1, \dots, c_k\})$ is input into the MLLM, which generates two outputs: (i) a natural language prompt $p$ that conforms to the semantic constraints of the selected theme $t$ and subject category $s$, and (ii) a structured description set $\{d_1, \dots, d_k\}$, where each $d_i$ specifies how the corresponding testpoint $c_i$ is realized in the prompt. Formally, this process can be expressed as:
\begin{equation}
(p, \{d_1, \dots, d_k\}) \sim \mathrm{MLLM}\Big(p, \{d_i\} \;\big|\; t, s, \{c_1, \dots, c_k\}\Big). 
\end{equation}
\subsubsection{T2I Model Evaluation} \label{Sec:bench_eval}

Given the generated images \( \{x_i\} \) for benchmark prompts \( \{p_i\} \), we evaluate each image using an MLLM. Specifically, the image \( x_i \), its corresponding prompt \( p_i \), and its testpoint descriptions \( \{d_{i,1}, \dots, d_{i,k}\} \) are provided as input. The MLLM evaluates each testpoint \( d_{i,j} \) in the context of \( x_i \), producing a binary score \( r_{i,j} \in \{0,1\} \) and a textual rationale \( e_{i,j} \) justifying the assessment. This can be formally represented as:
\begin{equation}
\begin{aligned}
(r_{i,1}, \dots, r_{i,k}, e_{i,1}, \dots, e_{i,k}) \sim \\\mathrm{MLLM}\Big(\{r_{i,j}, e_{i,j}\} \;\big|\; x_i, p_i, \{d_{i,1}, \dots, d_{i,k}\}\Big).
\end{aligned}
\end{equation}
This process ensures that the evaluation captures both the quantitative performance on each testpoint and the qualitative reasoning behind the assessment.

After obtaining the scores $r_{i,j}$ for each testpoint $d_{i,j}$ in all generated images, we aggregate them to compute scores of sub- and primary evaluation dimensions. Specifically, for each sub-dimension $c$, we define its score as the ratio of the number of times the model successfully satisfies the corresponding testpoint description to the total number of occurrences of that testpoint across the benchmark:
\begin{equation}
R_c = \frac{\sum_{i,j} \mathbf{1}\{d_{i,j} \in c \text{ and } r_{i,j}=1\}}{\sum_{i,j} \mathbf{1}\{d_{i,j} \in c\}},
\end{equation}
where $\mathbf{1}\{\cdot\}$ is the indicator function. The overall score for a primary dimension $C$ is then obtained by averaging the scores of all its sub-dimensions.
This procedure ensures that both fine-grained performance on sub-dimensions and broader performance on primary dimensions are captured.

\section{Experiments}

\subsection{Experimental Setup}
\textbf{Baselines.} We use FLUX.1-dev \cite{flux} as the base T2I model and UnifiedReward-Think \cite{unifiedreward-think} as the pairwise preference reward model (PPRM) in \ourmethod. As reward-maximization baselines, we train FLUX.1-dev with Flow-GRPO~\cite{flowgrpo} using each of HPSv2 \cite{hpsv2}, CLIP \cite{clip}, and UnifiedReward (UR) \cite{unifiedreward} as the pointwise reward.

\textbf{Training.} Using the pipeline in Fig.~\ref{fig:bench_pipeline}(a), we generate 5k training prompts, disjoint from the 600 \ourbench evaluation prompts. The same 5k prompts are used to train \ourmethod and all reward-maximization baselines.
Training is conducted on 64 H20 GPUs with 25 sampling steps, 8 rollouts per prompt from the same initial noise $x_T$, 4 gradient accumulation steps, and a learning rate of $1\times10^{-5}$. Following Flow-GRPO \cite{flowgrpo}, the SDE noise-scale hyperparameter is set to $a=0.7$. We host the PPRM as an inference server using vLLM~\cite{vllm}.

\textbf{Inference.} At inference, we generate $1024{\times}1024$ images with 30 sampling steps and a classifier-free guidance scale of 3.5, following the official FLUX.1-dev configuration. To ensure a fair comparison, all methods share identical prompts and initial noise at evaluation.

\textbf{Evaluation.} We evaluate semantic consistency on \ourbench, GenEval \cite{ghosh2023geneval}, and T2I-CompBench \cite{t2i-compbench}. For each test prompt, we generate four images and report the average score. Image quality is further evaluated with UnifiedReward \cite{unifiedreward}, ImageReward \cite{imagereward}, PickScore \cite{pickscore}, and Aesthetic \cite{aesthetics}. ImageReward, PickScore, and Aesthetic are disjoint from the training rewards and serve as the independent quality metrics.

\subsection{Main Results: \ourmethod vs.\ Reward-Maximization Baselines}
\textbf{Quantitative Results.} Tab.~\ref{tab:benchmark_results_pref-grpo} shows that \ourmethod delivers clear semantic gains on \ourbench. Relative to the strongest score-maximization baseline (UR), it improves the \textit{overall} score by 5.84\%, with especially large gains on challenging dimensions such as \textit{Text} (+12.69\%) and \textit{Logical Reasoning} (+12.04\%). This pattern is consistent with our motivation in Sec.~\ref{sec:adv_exp}: pairwise preferences provide a cleaner gradient signal than pointwise scores on dimensions that require fine-grained semantic grounding and compositional understanding.

The gains also generalize beyond our held-out evaluation on \ourbench. As shown in Tabs.~\ref{tab:geneval} and \ref{tab:t2i_comp}, \ourmethod consistently outperforms reward-maximization baselines on both GenEval and T2I-CompBench, indicating improved out-of-domain compositional reasoning, attribute binding, and spatial understanding. Moreover, Tab.~\ref{tab:ood_compare} shows that these semantic gains do not come at the expense of perceptual quality: \ourmethod achieves the best score on all four image-quality metrics. Overall, pairwise preference fitting improves semantic alignment without sacrificing image quality.

\begin{table*}[t]
\centering
\scriptsize
\setlength{\tabcolsep}{2.2pt}
\renewcommand{\arraystretch}{1.05}
\caption{\textbf{In-domain semantic consistency comparison on \ourbench.} FLUX.1-dev fine-tuned with Flow-GRPO using different reward models, versus our \ourmethod. Gemini-2.5-Pro is used as the evaluator. Best scores are in \textbf{bold}, second-best are \underline{underlined}.}
\vspace{-6pt}
\begin{tabularx}{\textwidth}{>{\raggedright\arraybackslash}p{2.2cm}YYYYYYYYYYY}
\toprule
\textbf{Model} & \textbf{Overall} & \textbf{Style} & \makecell{\textbf{World}\\\textbf{Know.}} & \textbf{Attr.} & \textbf{Action} & \textbf{Rel.} & \makecell{\textbf{Logic}\\\textbf{Reason.}} & \textbf{Grammar} & \textbf{Compound} & \textbf{Layout} & \textbf{Text} \\
\midrule
FLUX.1-dev        &61.30   & 83.90 & 88.92 & 67.84 & 62.17 & 67.26 & 30.91 & 60.96 & 47.04 & 71.83 & 32.18 \\
w/ HPSv2 &58.77   & 75.20 & 88.77 & 66.56 & 58.94 & 66.88 & 28.18 & 58.02 & 45.88 & 67.91 & 31.32 \\
w/ HPSv2\&CLIP  &61.81  & 84.92 & 88.98 & 68.44 & 62.54 & 68.10 & 31.01 & 59.36 & 50.60 & 71.07 & 33.07 \\
w/ UnifiedReward  &\underline{63.62}  & \underline{86.10} & \underline{89.72} & \underline{71.55} & \underline{63.69} & \underline{70.42} & \underline{32.05} & \underline{62.43} & \underline{52.32} & \underline{73.51} & \underline{34.44} \\
\midrule
\rowcolor[HTML]{E2F4E3} \textbf{w/ Pref-GRPO} & \textbf{69.46} & \textbf{88.40} & \textbf{90.35} & \textbf{75.00} & \textbf{69.77 } & \textbf{76.52} & \textbf{44.09} & \textbf{63.27} & \textbf{62.43} & \textbf{77.61} & \textbf{47.13} \\
\bottomrule
\end{tabularx}
\label{tab:benchmark_results_pref-grpo}

\end{table*}

\begin{figure*}[th]

    \centering
    \includegraphics[width=0.9\linewidth]{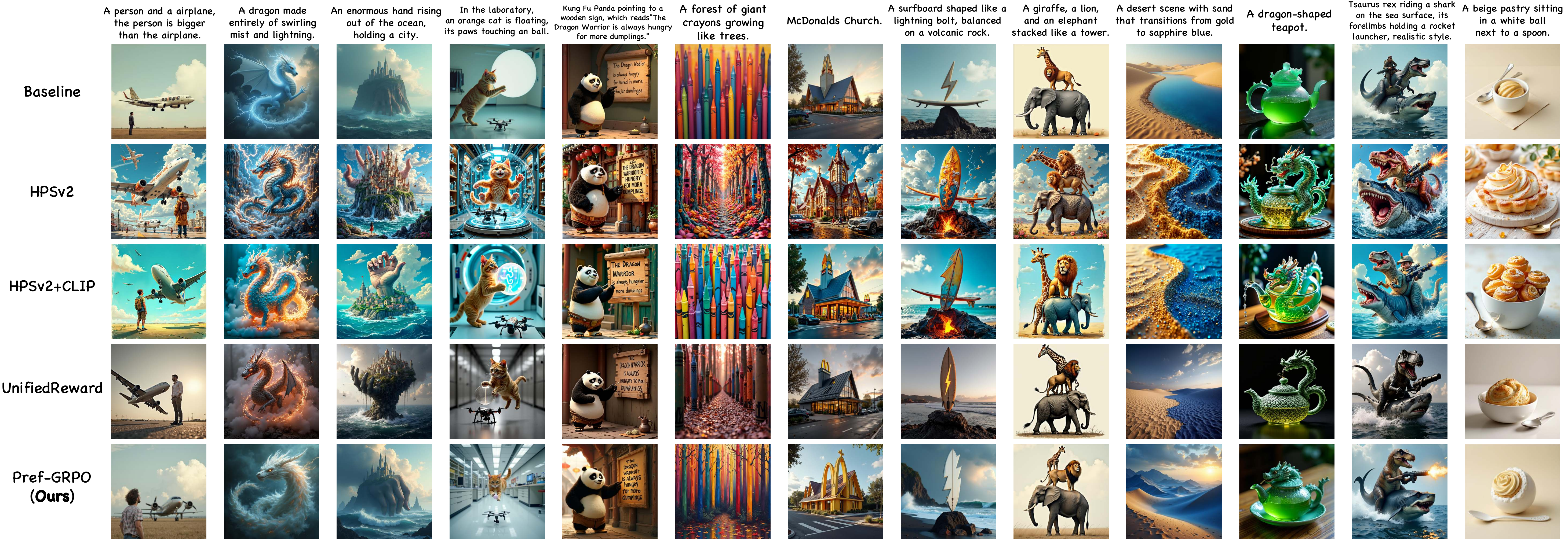}
    \vspace{-6pt}
    \caption{\textbf{Qualitative comparison.} Rows show FLUX.1-dev trained with HPSv2, HPSv2+CLIP, UnifiedReward, and \ourmethod. Pointwise-reward baselines exhibit reward-hacked artifacts (oversaturation for HPSv2, dark style for UnifiedReward), while \ourmethod preserves semantic faithfulness and the original FLUX.1-dev visual style.}
    \label{fig:qualitative}
\end{figure*}

\textbf{Qualitative Results.} Examples are shown in Fig. \ref{fig:qualitative}. Existing methods exhibit varying degrees of reward hacking: HPSv2-optimized images tend to be oversaturated, while UR-optimized images drift toward an unnaturally dark style. We also explore mitigating reward hacking by combining multiple reward scores, \ie summing HPSv2 and CLIP scores during training (third row in Fig. \ref{fig:qualitative}). While this partially reduces reward hacking, it does not eliminate it. In contrast, \ourmethod avoids these artifacts and faithfully renders the prompt semantics.

A detailed analysis of reward hacking under UnifiedReward and HPSv2 is provided in Sec.~\ref{sec:reward_hacking_analysis}.

\begin{table}[t]
\centering
\footnotesize
\setlength{\tabcolsep}{4pt}
\renewcommand{\arraystretch}{1.10}
\caption{\textbf{Image quality comparison.} FLUX.1-dev fine-tuned with Flow-GRPO using different reward models, versus our \ourmethod. ImageReward, PickScore, and Aesthetic are not used as training rewards and serve as independent quality metrics. Best results are in \textbf{bold}, second-best are \underline{underlined}.}
\vspace{-6pt}
\begin{tabularx}{\columnwidth}{lYYYY}
\toprule
\textbf{Model} & \makecell{\textbf{Unified}\\\textbf{Reward}} & \makecell{\textbf{Pick}\\\textbf{Score}} & \makecell{\textbf{Image}\\\textbf{Reward}} & \textbf{Aes.} \\
\midrule
FLUX.1-dev   & 3.04 & 22.42 & 1.27 &  6.13 \\

w/ HPSv2   & 3.09 & 22.62 & 1.34 &  6.20 \\
w/ HPSv2+CLIP & 3.08 & 22.61 & 1.30 &  6.25 \\
w/ UnifiedReward & \underline{3.14} & \underline{22.88} & \underline{1.38} &  \underline{6.31} \\
\midrule

\rowcolor[HTML]{E2F4E3} \textbf{w/ Pref-GRPO} & \textbf{3.26} & \textbf{23.02} & \textbf{1.44} & \textbf{6.52}\\

\bottomrule
\end{tabularx}
\label{tab:ood_compare}
\end{table}

\begin{table*}[t]
\centering
\scriptsize
\setlength{\tabcolsep}{3pt}
\renewcommand{\arraystretch}{1.08}
\caption{\textbf{Out-of-domain performance comparison on T2I-CompBench.} FLUX.1-dev fine-tuned with Flow-GRPO using different reward models, versus our \ourmethod. Best results are in \textbf{bold}, second-best are \underline{underlined}.}
\vspace{-6pt}
\begin{tabularx}{\textwidth}{lYYYYYYYYY}
\toprule
\textbf{Model} & \textbf{Overall} & \textbf{Color} & \textbf{Shape} & \textbf{Texture} & \makecell{\textbf{2D}\\\textbf{Spatial}} & \makecell{\textbf{3D}\\\textbf{Spatial}} & \textbf{Numeracy} & \makecell{\textbf{Non-}\\\textbf{Spatial}} & \textbf{Complex} \\
\midrule
FLUX.1-dev       & 48.17 & 77.34 & 48.32 & 62.66 & 28.01 & 40.04 & 61.88 & 30.67 & 36.49 \\
w/ HPSv2       & 46.77  & 78.17 & 51.55  &  66.13 & 22.06 & 33.75 &  56.34 &30.20  & 35.96 \\
w/ HPSv2+CLIP       & 49.18 & \underline{78.44} & 53.22 & 64.24 & 26.90 & 40.83 & 61.58 & 30.56 & 37.69 \\
w/ UnifiedReward       & \underline{50.20} & 78.32 & \underline{55.13} & \underline{67.44} & \underline{28.91} & \underline{40.04} & \underline{62.47} & \underline{30.88} & \underline{38.39} \\

\midrule
\rowcolor[HTML]{E2F4E3} \textbf{w/ Pref-GRPO} & \textbf{51.85} & \textbf{80.27} & \textbf{56.01} & \textbf{69.12} & \textbf{28.93} & \textbf{43.95} & \textbf{65.92} & \textbf{31.05} & \textbf{39.58} \\
\bottomrule
\end{tabularx}

\label{tab:t2i_comp}
\end{table*}

\subsection{Benchmarked T2I Models on \ourbench}
We further benchmark representative closed-source and open-source T2I models on \ourbench, spanning diffusion, MMDiT, autoregressive, and unified architectures. Closed-source models are accessed via their official APIs, and open-source models via their released checkpoints, both using the providers' default inference settings.

\textbf{Closed-source Models.} We benchmark GPT-4o~\cite{hurst2024gpt} (OpenAI), Imagen-4.0-Ultra~\cite{imagen} (Google), Seedream-3.0 and Seedream-4.0~\cite{gao2025seedream} (ByteDance), Nano Banana (Google), DALL-E-3 (OpenAI), FLUX-Pro-Ultra and Kontext-Max~\cite{flux} (Black Forest Labs), and Keling-Ketu (Kuaishou).

\textbf{Open-source Models.} We benchmark Qwen-Image~\cite{qwen_image}, HiDream-I1-Full~\cite{cai2025hidream}, Show-o2~\cite{show-o2}, SD-3.5-Large~\cite{sd}, SDXL~\cite{sd}, FLUX.1-dev~\cite{flux}, CogView4~\cite{cogview}, Hunyuan-DiT~\cite{li2024hunyuandit}, Playground 2.5~\cite{li2024playground}, Janus~\cite{janus}, Janus-Pro~\cite{janus-pro}, Janus-Flow~\cite{ma2024janusflow}, Emu3~\cite{wang2024emu3}, Bagel~\cite{deng2025emerging}, and BLIP3-o~\cite{blip3o}.

\subsection{Benchmarking Results on \ourbench} \label{sec:benchmarking_results}
\begin{table*}[t]
\centering
\scriptsize
\setlength{\tabcolsep}{3pt}
\renewcommand{\arraystretch}{1.12}
\caption{\textbf{Benchmarking results on \ourbench.} We report the overall score and primary-dimension breakdown for the evaluated T2I models. Best and second-best scores are computed separately within the closed- and open-source groups; they are marked in \textbf{bold} and \underline{underlined}, with top-3 per group annotated by the medal icons.}
\vspace{-6pt}
\begin{tabular*}{\textwidth}{@{\extracolsep{\fill}}lc@{\hskip 4pt}cccccccccc@{}}
\toprule
\textbf{Model} & \textbf{Overall} & \textbf{Style} & \makecell{\textbf{World}\\\textbf{Know.}} & \textbf{Attr.} & \textbf{Action} & \textbf{Rel.} & \makecell{\textbf{Logic}\\\textbf{Reason.}} & \textbf{Grammar} & \textbf{Compound} & \textbf{Layout} & \textbf{Text} \\
\midrule

\multicolumn{12}{c}{\textbf{Closed-source Models}} \\
\midrule

Keling-Ketu   & 65.93 & 92.27 & 86.62 & 71.66 & 68.73 & 70.94 & 43.75 & 71.26 & 60.81 & 77.23 & 16.03 \\
DALL-E-3    & 68.85  & 94.43 & 92.64 & 75.76 & 70.78 & 78.31 & 46.22 & 69.22 & 71.08 & 65.65 & 24.43 \\
FLUX-Pro-Ultra   & 70.46 & 90.99 & 91.30 & 76.79 & 71.39 & 78.05 & 41.46 & 68.18 & 68.17 & 80.60 & 37.64 \\
Seedream-3.0 & 78.41 & 98.19 & 94.90 & 84.62 & 83.14 & 80.18 & 51.83 & 60.30 & 72.32 & 88.74 & 69.86 \\
FLUX-Kontext-Max & 80.00 & 96.59 & 94.19 & 80.93 & 77.38 & 85.08 &61.36& 78.53 & 78.99 & 85.04 & 61.92 \\
Seedream-4.0 & 87.35 & 98.80 & 95.41 & 88.57 & 85.65 & 87.69 & 67.73 & 78.88 & 86.08 & 90.67 & \textbf{93.97} \\
\raisebox{-0.3em}{\includegraphics[height=1.2em]{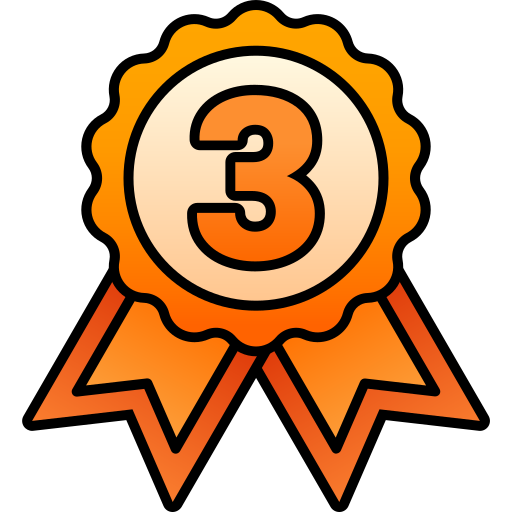}} Nano Banana & 87.29 & 98.59 & 96.20 & 87.99 & 87.36 & 92.47 & 73.41 & 83.82 & 88.34 & \underline{91.42} & 73.28 \\
\raisebox{-0.3em}{\includegraphics[height=1.2em]{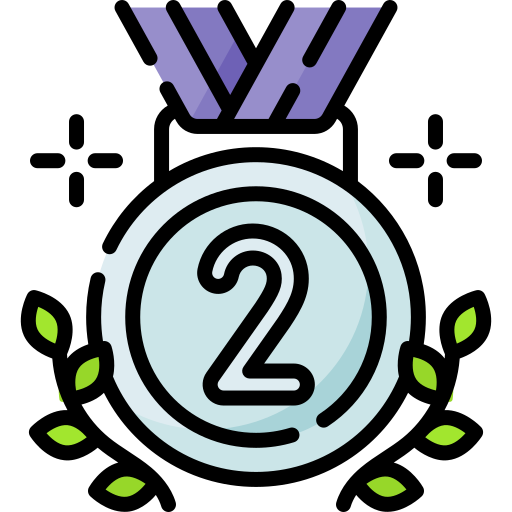}} Imagen-4.0-Ultra & \underline{91.65} & \textbf{99.10} & \underline{97.78} & \underline{92.09} & \textbf{92.10} & \underline{93.53} & \underline{80.45} & \underline{87.83} & \underline{91.37} & \textbf{92.91} & \underline{89.37} \\
\raisebox{-0.3em}{\includegraphics[height=1.2em]{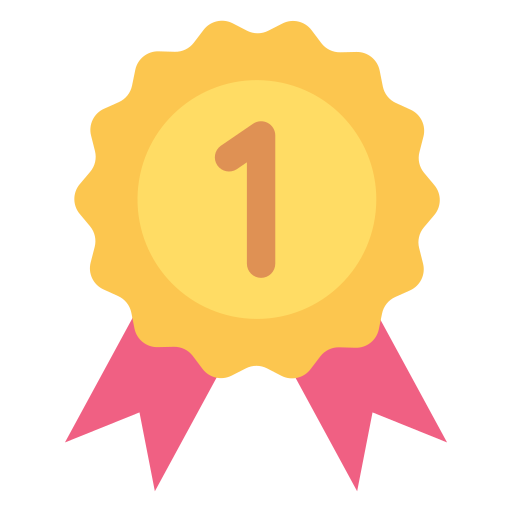}} \textbf{GPT-4o} & \textbf{92.48}  & \underline{98.98} & \textbf{98.22} & \textbf{94.01} & \underline{90.78} & \textbf{94.33} & \textbf{83.79} & \textbf{91.21} & \textbf{92.89} & 91.35 & 89.24 \\
\midrule

\multicolumn{12}{c}{\textbf{Open-source Models}} \\
\midrule
SDXL     & 40.22      & 87.45 & 72.28 & 44.66 & 35.10 & 46.37 & 10.34 & 48.48 & 26.68 & 30.80 & 0.00 \\
Playground 2.5 & 45.61 & 89.50 & 76.11 & 52.78 & 42.68 & 51.52 & 16.59 & 53.21 & 35.44 & 37.13 &  1.15 \\
Emu3         &45.42 & 87.50 & 76.42 & 50.11 & 40.40 & 48.60 & 19.32 & 50.67 & 36.21 & 43.84 &  1.15 \\
Janus-flow   &47.10  & 86.34 & 62.98 & 49.20 & 43.57 & 51.45 & 22.41 & 62.80 & 46.49 & 45.76 & 0.00 \\
Janus       &51.60   & 90.08 & 73.56 & 55.34 & 50.92 & 56.54 & 28.74 & 61.74 & 47.10 & 52.01 & 0.00 \\
Hunyuan-DiT       &51.38   & \underline{94.10} & 80.70 & 62.71 & 49.05 & 59.64 & 24.55 & 55.48 & 41.62 & 44.78 &  1.15 \\
CogView4    &56.00   & 80.80 & 81.96 & 63.14 & 59.51 & 60.91 & 27.95 & 54.81 & 44.97 & 69.03 & 16.95 \\
BLIP3-o    &  59.57   & 92.81 & 79.97 & 64.77 & 64.59 & 65.99 & 36.78 & \textbf{69.05} & 54.57 & 67.19 & 0.00 \\
FLUX.1-dev        &60.97   & 85.00 & 87.50 & 67.20 & 62.26 & 66.88 & 29.77 & 62.30 & 45.75 & 70.90 & 32.18 \\
Bagel      &  59.91  & 90.08 & 85.42 & 67.73 & 62.14 & 70.64 & 23.85 & \underline{65.85} & 56.86 & 76.56 & 0.00 \\
Janus-Pro   &61.36   & 90.40 & 86.55 & 68.59 & 63.88 & 69.54 & 35.68 & 64.04 & 60.18 & 72.76 & 2.01 \\
Show-o2    & 61.90   & 87.40 & 85.44 & 69.87 & 69.01 & 68.78 & 39.55 & 60.83 & \underline{63.79} & 73.13 & 1.15 \\
SD-3.5-Large   &62.89 & 88.60 & 89.72 & 68.80 & 61.98 & 67.51 & 32.05 & 59.89 & 58.38 & 67.72 & 34.20 \\
\raisebox{-0.3em}{\includegraphics[height=1.2em]{figures/bro_medal.png}}  Pref-GRPO &69.46   & 88.40 & 90.35 & \underline{75.00} & 69.77& \underline{76.52} & \underline{44.09} & 63.27 & 62.43 & \underline{77.61} & 47.13 \\
\raisebox{-0.3em}{\includegraphics[height=1.2em]{figures/medal.png}} HiDream-I1-Full     & \underline{71.36}  & 92.30 & \underline{93.67} & 73.40 & \underline{72.53} & 74.24 & 40.45 & 62.43 & 60.31 & \underline{77.61} & \underline{66.67} \\
\raisebox{-0.3em}{\includegraphics[height=1.2em]{figures/winner.png}} \textbf{Qwen-Image}  & \textbf{78.36}   & \textbf{94.70} & \textbf{94.15} & \textbf{87.93} & \textbf{82.60}& \textbf{80.08} & \textbf{51.59} & 60.96 & \textbf{72.94} & \textbf{86.57} & \textbf{72.13} \\
\bottomrule
\end{tabular*}
\label{tab:benchmark_results_main}
\end{table*}

\begin{table}[t]
\centering
\scriptsize
\setlength{\tabcolsep}{2pt}
\renewcommand{\arraystretch}{1.08}
\caption{\textbf{Out-of-domain performance comparison on GenEval.} FLUX.1-dev fine-tuned with Flow-GRPO using different reward models, versus our \ourmethod. Best results are in \textbf{bold}, second-best are \underline{underlined}.}
\vspace{-6pt}
\begin{tabularx}{\columnwidth}{lYYYYYYY}

\toprule
\textbf{Model} & \textbf{Overall} & \makecell{\textbf{Single}\\\textbf{Obj.}} & \makecell{\textbf{Two}\\\textbf{Obj.}} & \makecell{\textbf{Count.}} & \textbf{Color} & \textbf{Pos.} & \makecell{\textbf{Attr.}\\\textbf{Bind.}} \\
\midrule
FLUX.1-dev       & 62.92 & 97.81 & 79.55 & 71.56 & 77.66 & 18.50 & 42.25 \\
w/ HPSv2       & 59.31 & 97.43 & 75.00 & 62.81 & 73.67 & 21.00 & 34.75 \\

w/ HPSv2+CLIP      & 64.85  & 98.12  & 81.00 & 71.81 & 78.44 & 19.00 & 40.75 \\

w/ UnifiedReward       & \underline{67.28} & \underline{98.43} & \underline{82.57} & \underline{72.25} & \underline{79.72} & \underline{21.25} & \underline{49.50} \\

\midrule
\rowcolor[HTML]{E2F4E3} \textbf{w/ Pref-GRPO}       & \textbf{70.53} & \textbf{99.38} & \textbf{86.36} & \textbf{74.06} & \textbf{81.12} & \textbf{26.00} & \textbf{57.25} \\

\bottomrule

\end{tabularx}

\label{tab:geneval}
\end{table}

As shown in Tab. \ref{tab:benchmark_results_main}, \ourbench reveals a clear yet nuanced performance hierarchy. \textbf{Closed-source models} remain the strongest overall, with GPT-4o \cite{hurst2024gpt} and Imagen-4.0-Ultra \cite{imagen} leading on 7 of 10 primary dimensions. Their advantage is most pronounced on dimensions that require deeper semantic grounding (e.g., \emph{Logical Reasoning}, \emph{Relationship}, \emph{Compound}, and \emph{Text}) rather than low-level visual quality. By contrast, dimensions such as \emph{Style} and \emph{World Knowledge} are already saturated across many models (most score above 0.8), indicating they are no longer the main bottlenecks for current T2I systems. \emph{This contrast illustrates why coarse overall scores are insufficient: the real gap between strong and weak models lies in higher-order semantic composition, not in basic visual plausibility.}

\textbf{Open-source models} are closing the gap but remain less balanced across dimensions. Qwen-Image \cite{qwen_image} is the strongest open-source model overall, and HiDream-I1-Full \cite{cai2025hidream} is competitive on \emph{Action}, \emph{Layout}, and \emph{Attribute}. Compared with leading closed-source systems, however, open-source models show larger variance across the 10 primary dimensions, with noticeably weaker performance on \emph{Logical Reasoning}, \emph{Grammar}, and \emph{Compound}. This indicates the remaining gap is not primarily one of image quality, but of fine-grained instruction grounding and coherently satisfying multiple constraints within a single prompt. \emph{\ourbench thus serves not only as a leaderboard, but also as a diagnostic benchmark that exposes where current T2I models still fail.}

Tab. \ref{tab:benchmark_results_finegrained} further sharpens this diagnosis by exposing the sub-dimension breakdown hidden beneath the primary-dimension averages. Among \textbf{closed-source models}, GPT-4o leads on semantically fine-grained sub-dimensions such as \emph{expression}, \emph{non-contact interaction}, \emph{Grammar}, and \emph{Logical Reasoning}, whereas Imagen-4.0-Ultra leads on structure-sensitive sub-dimensions including \emph{quantity}, \emph{contact interaction}, \emph{composition}, and \emph{3D spatial arrangement} (Layout). Even among top-tier systems, strengths are not uniform: some models are stronger on abstract semantic interpretation, others on preserving spatial and compositional structure.
The fine-grained breakdown is even more informative for \textbf{open-source models}. Qwen-Image shows the most balanced profile and leads in a majority of sub-dimensions among open-source systems, indicating broad rather than narrow improvements. HiDream-I1-Full is competitive on several geometry- and action-related sub-dimensions, but remains visibly weaker on reasoning-intensive dimensions (\emph{Logical Reasoning}, \emph{Grammar}, and \emph{Compound}). Crucially, models with similar primary-dimension averages can still differ substantially in the specific sub-dimensions they master, which \ourbench makes explicit.

\begin{table*}[t]
\centering
\tiny
\setlength{\tabcolsep}{1.45pt}
\renewcommand{\arraystretch}{1.12}
\caption{\textbf{Fine-grained benchmarking results of the models reported in Tab.~\ref{tab:benchmark_results_main}.} We report sub-dimension breakdowns for overlapping models with available fine-grained annotations. Best scores are in \textbf{bold}, second-best in \underline{underlined}, computed separately for closed- and open-source models.}
\vspace{-6pt}
\resizebox{\textwidth}{!}{%
\begin{tabular}{lc c|cc c c c c|cc c c c c|cc c c|cc|cc c|cc|cc}
\toprule
\textbf{Model} & \textbf{Style} & \makecell{\textbf{World}\\\textbf{Know.}}
& \multicolumn{6}{c|}{\textbf{Attribute}}
& \multicolumn{6}{c|}{\textbf{Action}}
& \multicolumn{4}{c|}{\textbf{Relationship}}
& \multicolumn{2}{c|}{\textbf{Compound}}
& \multicolumn{3}{c|}{\textbf{Grammar}}
& \multicolumn{2}{c|}{\textbf{Layout}}
& \makecell{\textbf{Logic}\\\textbf{Reason.}}
& \textbf{Text} \\
\cmidrule(lr){4-9}
\cmidrule(lr){10-15}
\cmidrule(lr){16-19}
\cmidrule(lr){20-21}
\cmidrule(lr){22-24}
\cmidrule(lr){25-26}
& & &
\textbf{Quant.} & \makecell{\textbf{Express.}} & \makecell{\textbf{Materi.}} & \textbf{Size} & \textbf{Shape} & \textbf{Color}
& \textbf{Hand} & \makecell{\textbf{Full}\\\textbf{Body}} & \textbf{Animal} & \makecell{\textbf{Non}\\\textbf{Contact}} & \textbf{Contact} & \textbf{State}
& \textbf{Compos.} & \textbf{Sim.} & \textbf{Inclus.} & \textbf{Compare.}
& \makecell{\textbf{Imagin.}} & \makecell{\textbf{Feat.}\\\textbf{Match.}}
& \makecell{\textbf{Pron.}\\\textbf{Ref.}} & \textbf{Consist.} & \textbf{Neg.}
& \textbf{2D} & \textbf{3D}
& & \\
\midrule
\multicolumn{28}{c}{\textbf{Closed-source Models}} \\
\midrule
Keling-Ketu & 92.25 & 87.08 & 74.29 & 56.77 & 78.67 & 74.83 & 53.75 & 89.66 & 53.85 & 72.28 & 71.32 & 70.77 & 59.28 & 75.94 & 68.14 & 69.27 & 72.13 & 69.29 & 66.15 & 53.03 & 74.91 & 64.19 & 66.80 & 77.61 & 71.43 & 45.60 & 16.03 \\
DALL-E-3 & 94.43 & 92.64 & 60.14 & 63.16 & 87.20 & 84.72 & 66.25 & 91.60 & 60.78 & 76.67 & 77.94 & 68.72 & 63.19 & 76.19 & 82.99 & 71.51 & 85.47 & 66.93 & 78.01 & 63.95 & 76.34 & 72.09 & 59.45 & 54.78 & 77.25 & 46.22 & 24.43 \\
FLUX-Pro-Ultra & 90.99 & 91.30 & 72.92 & 60.65 & 79.25 & 75.00 & 78.12 & 98.33 & 58.97 & 69.02 & 76.47 & 78.06 & 65.48 & 77.83 & 81.08 & 74.44 & 80.98 & 71.88 & 77.30 & 58.85 & 83.46 & 65.74 & 54.23 & 81.25 & 79.92 & 41.46 & 37.64 \\
Seedream-3.0 & 98.19 & 94.90 & 79.02 & 81.94 & 89.62 & 83.80 & 77.22 & 96.67 & 75.97 & 89.56 & 86.03 & 75.38 & 81.93 & 89.10 & 81.57 & 74.16 & 83.61 & 80.47 & 76.92 & 67.62 & 77.94 & 68.40 & 35.14 & 88.15 & 89.35 & 51.83 & 69.86 \\
FLUX-Kontext-Max & 96.59 & 94.19 & 75.69 & 74.32 & 82.55 & 86.81 & 74.38 & 94.17 & 67.95 & 83.15 & 77.94 & 77.04 & 70.83 & 84.43 & 87.50 & 78.89 & 90.00 & 81.25 & 83.93 & 73.96 & 84.23 & 78.70 & 72.69 & 86.74 & 83.33 & 61.36 & 61.92 \\
Seedream-4.0 & 98.80 & 95.41 & 86.81 & \underline{85.90} & \textbf{97.17} & 84.03 & 76.88 & \textbf{100.00} & 77.56 & 87.50 & \underline{88.24} & 80.10 & 83.93 & 94.81 & 88.18 & 80.56 & 94.02 & 87.50 & 88.27 & 83.85 & 84.93 & 79.17 & 72.31 & 90.81 & \underline{90.53} & 67.73 & \textbf{93.97} \\
\raisebox{-0.3em}{\includegraphics[height=1.2em]{figures/bro_medal.png}} Nano Banana & 98.59 & 96.20 & 86.43 & 80.77 & 88.46 & \textbf{95.83} & 80.77 & 98.33 & 80.13 & \textbf{93.48} & \underline{88.24} & 83.67 & 80.95 & \underline{95.28} & 93.49 & 86.67 & 94.02 & \textbf{96.09} & 90.21 & 86.46 & 90.44 & 83.33 & 77.31 & \underline{93.01} & 89.77 & 73.41 & 73.28 \\
\raisebox{-0.3em}{\includegraphics[height=1.2em]{figures/medal.png}} Imagen-4.0-Ultra & \textbf{99.10} & \underline{97.78} & \textbf{94.44} & 80.77 & \underline{95.28} & \underline{94.44} & \underline{88.75} & \textbf{100.00} & \textbf{89.74} & \underline{93.41} & \textbf{93.38} & \underline{88.78} & \underline{87.50} & \textbf{98.58} & \underline{96.28} & \underline{87.78} & \textbf{96.20} & 91.41 & \underline{92.86} & \underline{89.84} & \textbf{91.91} & \underline{90.28} & \underline{81.54} & \textbf{93.75} & \textbf{92.05} & \underline{80.45} & \underline{89.37} \\
\raisebox{-0.3em}{\includegraphics[height=1.2em]{figures/winner.png}} \textbf{GPT-4o} & \underline{98.98} & \textbf{98.22} & \underline{89.29} & \textbf{96.00} & 94.66 & 92.96 & \textbf{92.50} & \underline{99.17} & \underline{88.46} & 93.33 & 87.88 & \textbf{92.02} & \textbf{89.16} & 92.31 & \textbf{96.58} & \textbf{91.11} & \underline{94.89} & \underline{92.97} & \textbf{94.07} & \textbf{91.67} & \underline{91.04} & \textbf{93.06} & \textbf{89.75} & 92.16 & \underline{90.53} & \textbf{83.79} & 89.24 \\
\midrule
\multicolumn{28}{c}{\textbf{Open-source Models}} \\
\midrule
SDXL & 87.45 & 72.28 & 41.67 & 25.00 & 54.90 & 44.85 & 36.11 & 68.52 & 19.74 & 38.10 & 45.31 & 26.74 & 24.34 & 52.40 & 55.38 & 41.22 & 38.75 & 43.33 & 33.75 & 19.94 & 54.58 & 41.67 & 47.46 & 25.00 & 36.40 & 10.34 & 0.00 \\
Playground 2.5 & 89.78 & 75.80 & 60.61 & 43.59 & 58.33 & 45.59 & 39.58 & 81.48 & 29.61 & 54.17 & 54.69 & 37.21 & 28.29 & 57.21 & 63.46 & 51.35 & 48.75 & 40.00 & 44.06 & 28.27 & 62.92 & 51.11 & 49.58 & 33.18 & 39.47 & 16.09 & 0.00 \\
Emu3 & 87.50 & 76.42 & 42.36 & 45.51 & 52.83 & 40.28 & 46.25 & 77.50 & 23.08 & 49.46 & 54.41 & 34.69 & 29.17 & 50.47 & 55.41 & 44.44 & 46.74 & 41.41 & 41.33 & 30.99 & 58.09 & 49.07 & 44.23 & 42.28 & 45.45 & 19.32 & 1.15 \\
Janus-flow & 86.34 & 62.98 & 43.18 & 30.77 & 55.39 & 57.35 & 33.33 & 82.41 & 22.37 & 48.81 & 57.81 & 38.95 & 36.84 & 54.81 & 62.69 & 36.49 & 53.75 & 42.50 & 60.00 & 33.63 & 70.00 & 51.11 & \textbf{64.41} & 46.82 & 44.74 & 22.41 & 0.00 \\
Janus & 90.08 & 73.56 & 35.61 & 37.82 & 60.29 & 66.18 & 48.61 & 90.74 & 31.58 & 52.38 & 62.50 & 50.00 & 39.47 & 65.87 & 58.85 & 52.70 & 61.25 & 50.00 & 59.38 & 35.42 & 70.00 & 52.22 & 60.59 & 51.82 & 52.19 & 28.74 & 0.00 \\
Hunyuan-DiT & \underline{94.10} & 80.70 & 67.36 & 44.23 & 71.70 & 61.81 & 47.50 & 86.67 & 35.90 & 54.89 & 54.41 & 46.94 & 35.71 & 62.74 & 60.14 & 64.44 & 60.33 & 50.78 & 46.68 & 36.46 & 62.87 & 57.87 & 45.77 & 39.34 & 50.38 & 24.55 & 1.15 \\
CogView4 & 80.80 & 81.96 & 70.83 & 46.79 & 55.66 & 68.75 & 58.75 & 87.50 & 57.69 & 59.78 & 69.85 & 52.55 & 53.57 & 65.09 & 58.11 & 60.00 & 66.30 & 60.94 & 49.23 & 40.62 & 69.49 & 54.17 & 40.00 & 76.84 & 60.98 & 27.95 & 16.95 \\
BLIP3-o & 92.81 & 79.97 & 48.48 & 60.26 & 66.67 & 76.47 & 56.94 & 83.33 & 57.24 & 71.43 & 71.09 & 63.95 & 50.66 & 71.15 & 70.77 & 57.43 & 66.25 & 65.83 & 64.06 & 45.54 & \underline{81.67} & 61.11 & \underline{62.29} & 69.55 & 64.91 & 36.78 & 0.00 \\
FLUX.1-dev & 85.00 & 87.50 & 71.53 & 51.92 & 58.96 & 74.31 & \underline{65.62} & 90.00 & 50.00 & 69.02 & 69.12 & 60.20 & 61.90 & 63.21 & 66.89 & 65.56 & 72.83 & 60.16 & 46.17 & 45.31 & 76.47 & 61.57 & 48.08 & 74.63 & 67.05 & 29.77 & 32.18 \\
Bagel & 90.08 & 85.42 & 56.82 & 50.00 & 73.53 & 77.94 & 59.03 & 94.44 & 51.32 & 64.88 & 67.19 & 64.53 & 56.58 & 66.83 & \underline{77.31} & \underline{68.92} & 70.00 & 59.17 & 67.50 & 46.73 & 74.17 & 64.44 & 58.47 & 77.73 & \underline{75.44} & 23.85 & 0.00 \\
Janus-Pro & 90.40 & 86.55 & 56.25 & 57.69 & \underline{74.06} & 73.61 & 61.88 & 90.83 & 47.44 & 65.22 & 72.79 & 60.71 & 59.52 & 75.47 & 76.01 & 58.33 & 73.91 & 64.06 & 67.35 & 52.86 & 76.10 & \underline{64.81} & 50.77 & 74.63 & 70.83 & 35.68 & 2.01 \\
Show-o2 & 87.40 & 85.44 & 59.03 & \underline{64.10} & 70.75 & 74.31 & 61.25 & 95.00 & 54.49 & 75.00 & \underline{75.00} & 72.45 & 50.60 & \underline{82.08} & 76.35 & 60.56 & 71.20 & 59.38 & 66.84 & \underline{60.68} & 77.57 & 63.43 & 41.15 & 75.37 & 70.83 & 39.55 & 1.15 \\
\raisebox{-0.3em}{\includegraphics[height=1.2em]{figures/bro_medal.png}} SD-3.5-Large & 88.60 & 89.72 & 69.44 & 51.28 & 70.28 & 70.83 & 64.38 & 91.67 & 57.69 & 63.04 & 62.50 & 59.69 & 58.93 & 68.40 & 73.99 & 65.00 & 66.30 & 57.81 & \underline{68.37} & 48.18 & 77.21 & 60.19 & 41.54 & 70.96 & 64.39 & 32.05 & 34.20 \\
\raisebox{-0.3em}{\includegraphics[height=1.2em]{figures/medal.png}} HiDream-I1-Full & 92.30 & \underline{93.67} & \underline{73.61} & 61.54 & 72.17 & \underline{79.17} & 62.50 & \underline{98.33} & \underline{60.90} & \underline{76.09} & 74.26 & \underline{73.98} & \underline{68.45} & 78.77 & 76.69 & 67.78 & \underline{78.26} & \underline{71.88} & 61.99 & 58.59 & 81.62 & 63.89 & 41.15 & \underline{82.72} & 72.35 & \underline{40.45} & \underline{66.67} \\
\raisebox{-0.3em}{\includegraphics[height=1.2em]{figures/winner.png}} \textbf{Qwen-Image} & \textbf{94.70} & \textbf{94.15} & \textbf{84.03} & \textbf{85.26} & \textbf{91.98} & \textbf{86.11} & \textbf{81.88} & \textbf{99.17} & \textbf{78.21} & \textbf{86.96} & \textbf{86.76} & \textbf{77.55} & \textbf{76.79} & \textbf{88.68} & \textbf{82.09} & \textbf{71.11} & \textbf{86.96} & \textbf{78.12} & \textbf{73.21} & \textbf{72.66} & \textbf{84.93} & \textbf{70.37} & 28.08 & \textbf{87.13} & \textbf{85.98} & \textbf{51.59} & \textbf{72.13} \\
\bottomrule
\end{tabular}
}
\label{tab:benchmark_results_finegrained}
\end{table*}

\section{Analyses and Ablations}
\begin{figure}[ht]

    \centering
    \includegraphics[width=1\linewidth]{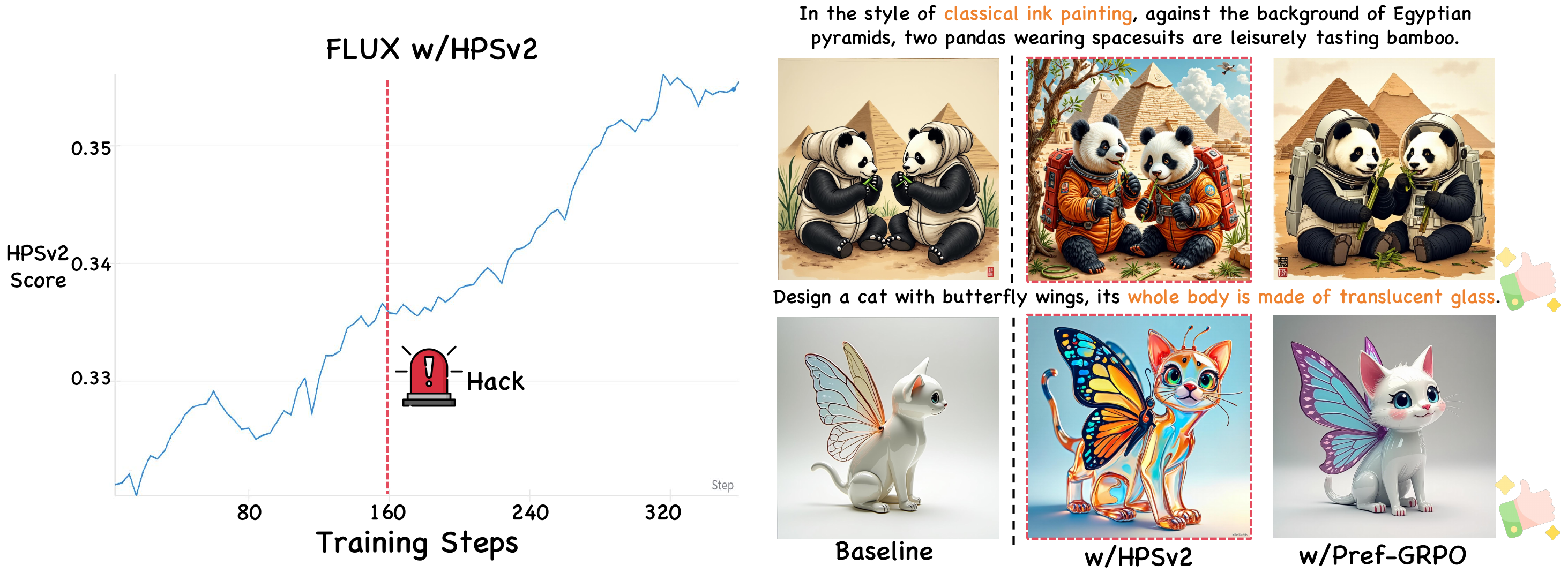}
    \vspace{-6pt}
    \caption{\textbf{Reward hacking under HPSv2.} HPSv2 scores continue to rise, but image quality collapses around step 160, manifesting as severe over-saturation.}
    \label{fig:hps_hacking}
\end{figure}

\subsection{Reward Hacking in Pointwise GRPO} \label{sec:reward_hacking_analysis}
We analyze reward hacking under two representative pointwise reward models, UnifiedReward~\cite{unifiedreward} and HPSv2~\cite{hpsv2}. For UnifiedReward, Fig.~\ref{fig:reward_hack} shows that reward scores rise rapidly during training, yet the generated images gradually drift toward an unnatural dark style and perceptual quality degrades. For HPSv2, the same failure mode emerges even more abruptly in Fig.~\ref{fig:hps_hacking}: the reward keeps rising, while image quality starts to collapse at around step 160, accompanied by severe over-saturation. Taken together, these results reveal a consistent mismatch between reward growth and real generation quality under reward score maximization.

This mismatch is consistent with the \textit{illusory advantage} analysis in Sec.~\ref{sec:adv_exp}. Once pointwise reward scores within a rollout group become highly compressed, GRPO amplifies tiny score gaps into large normalized advantages, causing the policy to over-optimize for spurious reward cues rather than genuine improvements in alignment or aesthetics. HPSv2 exhibits reward hacking earlier than UnifiedReward, consistent with its smaller intra-group score variance in Fig. \ref{fig:teaser}(a), which intensifies the amplification effect. By contrast, \ourmethod optimizes relative preferences rather than absolute-score increments, leading to substantially more stable training behavior.

Several complementary observations support this mitigation. First, Fig.~\ref{fig:teaser}(a) shows that \ourmethod maintains a larger intra-group reward variance, directly suppressing the illusory-advantage amplification mechanism. Second, Figs.~\ref{fig:reward_hack} and \ref{fig:hps_hacking} show that, unlike pointwise-score GRPO, \ourmethod does not drift toward visually degraded solutions during prolonged training. Third, the qualitative comparison in Fig.~\ref{fig:qualitative} shows that \ourmethod preserves the original FLUX.1-dev visual style and does not collapse to reward-hacked artifacts. Finally, as we show next in Sec. \ref{sec:point_winrate}, a controlled point-score-derived win-rate variant further confirms that replacing absolute-score maximization with a relative-comparison objective is itself the key stabilizing factor.

\subsection{Point-Score vs. Pairwise-Preference Win Rates} \label{sec:point_winrate}
To disentangle the effect of the optimization objective from that of the reward model itself, we introduce a controlled variant based on \emph{point score-derived win rates}. Specifically, we first compute UnifiedReward scores for all images in a rollout group, then convert these scores into pairwise win rates by comparing every image pair, and finally use the resulting win rates as rewards for training. This keeps the underlying reward model unchanged while replacing reward score maximization with a relative-comparison objective.

\begin{figure}[t]

    \centering
    \includegraphics[width=0.8\linewidth]{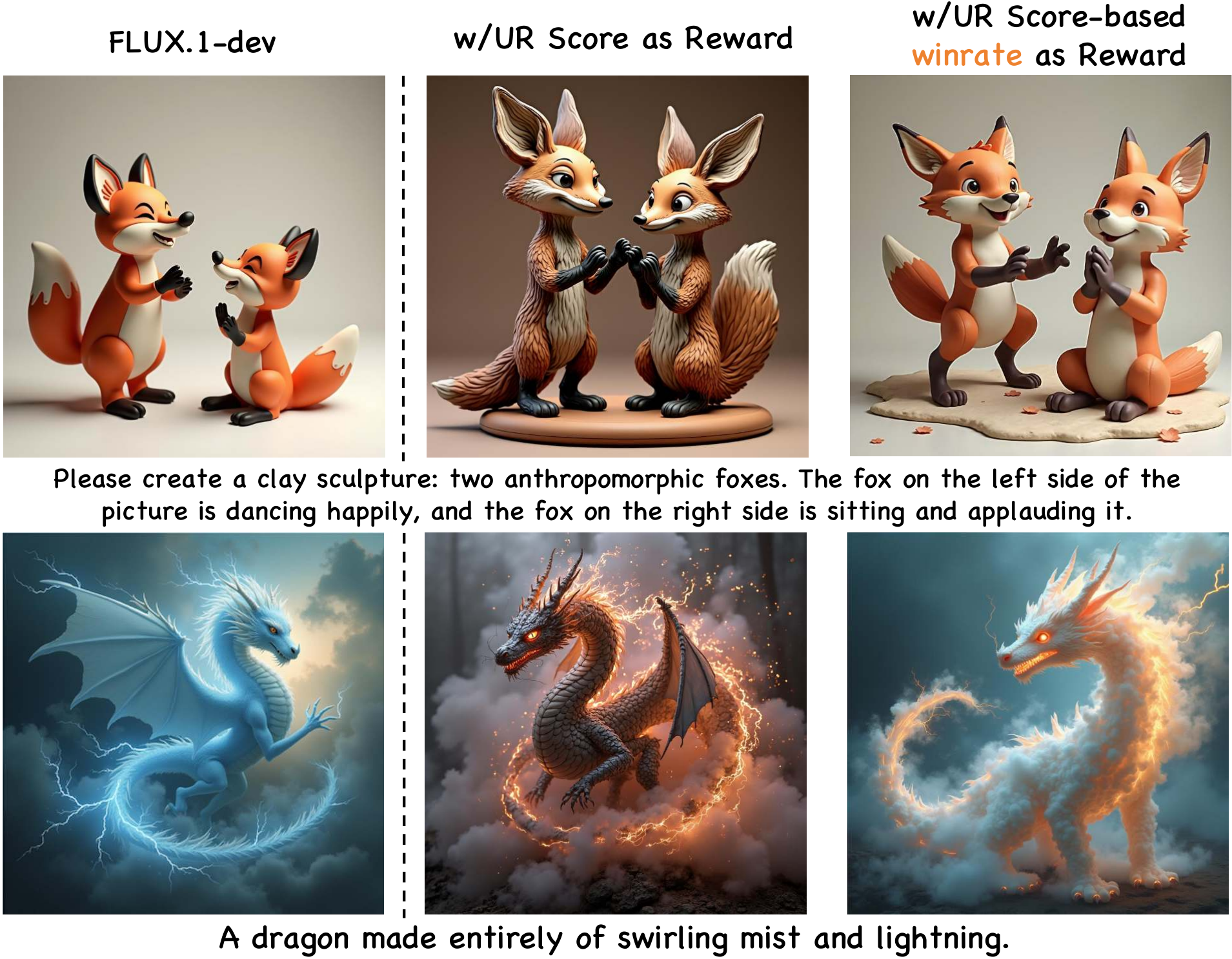}
    \vspace{-6pt}
    \caption{\textbf{Point-score-derived win-rate training alleviates reward hacking.} Converting UnifiedReward scores into pairwise win rates substantially mitigates the dark-style drift observed under direct score maximization, supporting relative preferences (not the reward model itself) as the key stabilizing factor.}
    \label{fig:supp_point_winrate}
    
\end{figure}

This ablation isolates the source of the stability gain. If the improvement of \ourmethod came solely from using a stronger pairwise reward model, then converting UnifiedReward scores into win rates should offer little benefit. However, Fig.~\ref{fig:supp_point_winrate} shows the opposite: once the optimization target is changed from reward score maximization to win-rate fitting, the dark-style reward hacking observed under UnifiedReward is already substantially alleviated. Quantitative results in Tab.~\ref{tab:point_winrate} further confirm this trend: training with point-score-derived win rates consistently outperforms reward score maximization, and \ourmethod, which uses native pairwise preference rewards, achieves the best overall performance.

These results support two conclusions. First, a significant part of the instability in prior GRPO methods arises from the \emph{objective formulation}, not merely from imperfections in the reward model. Second, pairwise preference rewards outperform score-derived win rates, indicating that native pairwise supervision provides cleaner and more faithful relative signals than rankings derived from compressed pointwise scores. In short, \ourmethod benefits from both components: relative-preference optimization is the key stabilizing factor, strengthened further by a dedicated pairwise reward model. A practical corollary is that practitioners with only scalar rewards can adopt win-rate fitting as a drop-in stability improvement.

\begin{table}[t]
\centering
\scriptsize
\setlength{\tabcolsep}{2pt}
\renewcommand{\arraystretch}{1.08}
\caption{\textbf{Point-score-derived win rate vs. pairwise preference win rate.} Compared to direct score maximization (``w/ UR (score)''), converting UnifiedReward scores into pairwise win rates (``w/ UR (score win rate)'') already improves stability; \ourmethod further benefits from a native pairwise preference reward. Best results are in \textbf{bold}, second-best are \underline{underlined}.}
\vspace{-6pt}
\begin{tabularx}{\columnwidth}{lYYYYYY}
\toprule
\textbf{Model} & \textbf{UniGen} & \textbf{GenEval} & \makecell{\textbf{Unified}\\\textbf{Reward}} & \makecell{\textbf{Pick}\\\textbf{Score}} & \makecell{\textbf{Image}\\\textbf{Reward}} & \textbf{Aes.} \\
\midrule
FLUX.1-dev & 61.30 & 62.92 & 3.04 & 22.42 & 1.27 & 6.13 \\
w/ UR (score) & 63.62 & 67.28 & 3.14 & 22.88 & 1.38 & 6.31 \\
\makecell[l]{w/ UR (score winrate)} & \underline{64.32} & \underline{68.13} & \underline{3.20} & \underline{22.91} & \underline{1.39} & \underline{6.35} \\
\rowcolor[HTML]{E2F4E3} \textbf{w/ Pref-GRPO} & \textbf{69.46} & \textbf{70.53} & \textbf{3.26} & \textbf{23.02} & \textbf{1.44} & \textbf{6.52} \\
\bottomrule
\end{tabularx}
\label{tab:point_winrate}
\end{table}

\subsection{Wall-Clock Overhead of Pairwise Rewards}
Although pairwise rewards require $O(G^2)$ comparisons within each rollout group of size $G$, the practical question is whether this additional computation becomes a prohibitive bottleneck during training. To answer this, we measure wall-clock reward computation time when training FLUX.1-dev on $4\times$ H800 GPUs, with the reward inference service deployed on another 4 H800 GPUs via vLLM~\cite{vllm}; this isolates per-step cost from the multi-node overhead of the 64-GPU main setup. We compare two settings across different group sizes: (i) point-score rewards from UnifiedReward, and (ii) pairwise preference win-rate rewards used in \ourmethod.

\begin{table}[t]
  \centering
  \footnotesize
  \setlength{\tabcolsep}{3pt}
  \renewcommand{\arraystretch}{1.05}
  \caption{\textbf{Wall-clock reward computing time per training step}. ``Requests/step'' counts the number of reward-model forward computations invoked by each scheme in a training step.}
  \vspace{-6pt}
  \label{tab:reward_wallclock}
  \begin{tabular*}{\columnwidth}{@{\extracolsep{\fill}}lccc@{}}
    \toprule
    \textbf{Reward type} & \makecell{\textbf{Rollouts}\\\textbf{/ group}} & \makecell{\textbf{Requests}\\\textbf{/ step}} & \makecell{\textbf{Time}\\\textbf{/ step}} \\
    \midrule
    \multirow{2}{*}{\makecell[l]{Point score\\(UnifiedReward)}} & 8  & 32  & 3\,s \\
                                                                 & 16 & 64  & 5\,s \\
    \midrule
    \multirow{3}{*}{\makecell[l]{Pairwise winrate\\(Ours)}}      & 8  & 112 & 7\,s \\
                                                                 & 12 & 264 & 14\,s \\
                                                                 & 16 & 480 & 22\,s \\
    \bottomrule
  \end{tabular*}
\end{table}

Tab.~\ref{tab:reward_wallclock} shows that the added cost is moderate. Under the default setting $G{=}8$, pairwise win-rate rewards increase the per-step reward time from 3\,s to 7\,s, an additional 4\,s of overhead. At $G{=}16$, the number of reward-model requests grows from 64 to 480 ($7.5\times$), yet wall-clock reward time grows only from 5\,s to 22\,s ($3.1\times$), far sublinear in the number of pairwise comparisons. Pairwise supervision is therefore effectively amortized by parallel reward serving.

\emph{The main concern with pairwise rewards is theoretical complexity, not practical infeasibility.} In our setting, $G{=}8$ already yields strong empirical gains (Tab.~\ref{tab:ood_compare}, Figs.~\ref{fig:reward_hack}, \ref{fig:qualitative}), so the added reward latency remains modest relative to the training benefit. Overall, \ourmethod achieves a favorable computation--performance trade-off: substantially improved generation quality and mitigated reward hacking, with only limited wall-clock overhead.
\begin{figure}[t]

    \centering
    \includegraphics[width=1\linewidth]{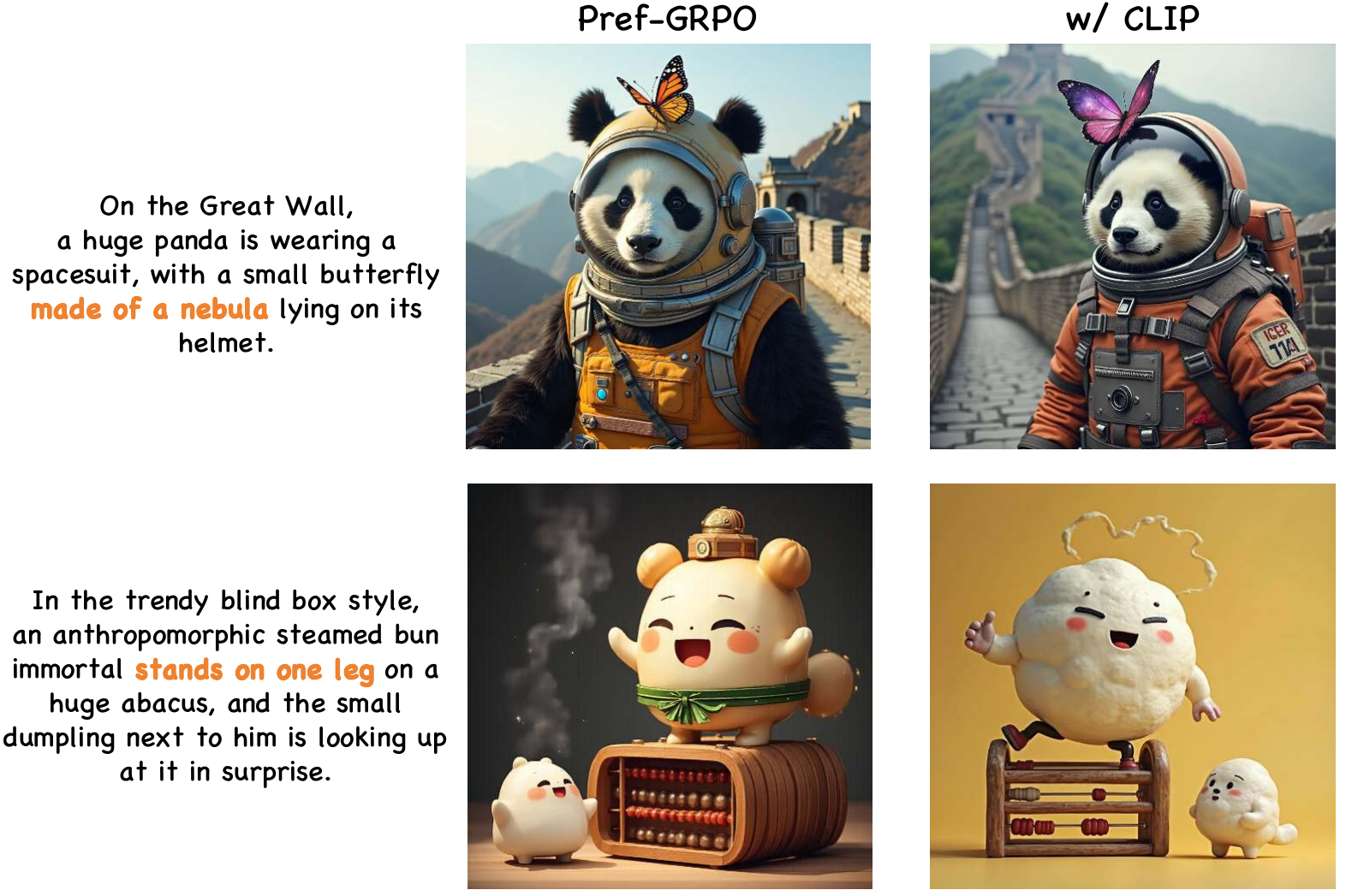}
    \vspace{-6pt}
    \caption{\textbf{Joint optimization of \ourmethod with a CLIP score reward.} Adding CLIP improves semantic consistency while slightly reducing perceptual quality, but does not trigger reward hacking. The pairwise preference fitting acts as a stabilizing regularizer for the auxiliary score signal.}
    \label{fig:pref_clip}
\end{figure}

\begin{table}[t]
\centering
\scriptsize
\setlength{\tabcolsep}{2pt}
\renewcommand{\arraystretch}{1.08}
\caption{\textbf{Joint optimization of \ourmethod and CLIP-based reward score maximization}. Best results are in \textbf{bold}.}
\vspace{-6pt}
\begin{tabular*}{\columnwidth}{@{\extracolsep{\fill}}lcccccc@{}}
\toprule
\textbf{Model} & \textbf{UniGen} & \textbf{GenEval} & \makecell{\textbf{Unified}\\\textbf{Reward}} & \makecell{\textbf{Pick}\\\textbf{Score}} & \makecell{\textbf{Image}\\\textbf{Reward}} & \textbf{Aes.} \\
\midrule
Pref-GRPO & 69.46 & 70.53 & \textbf{3.26} & \textbf{23.02} & \textbf{1.44} & \textbf{6.52}\\
\makecell[l]{{w/ CLIP}} & \textbf{70.02} & \textbf{71.26} & 3.18 & 22.86 & 1.41 & 6.44\\
\bottomrule
\end{tabular*}
\label{tab:joint_optimization}
\end{table}

\subsection{Joint Optimization with Auxiliary Score Rewards}
To examine whether score-based rewards can still provide useful auxiliary supervision, we add a CLIP \cite{clip} reward on top of \ourmethod and perform joint optimization. Here CLIP serves as a score-based reward signal that focuses more on semantic consistency between the prompt and the generated image. This experiment tests whether pairwise preference fitting can serve as a stabilizing objective while score-based rewards provide complementary semantic guidance.

As shown in Tab.~\ref{tab:joint_optimization}, joint optimization improves semantic consistency on \ourbench and GenEval, but slightly reduces image-quality-oriented metrics relative to vanilla \ourmethod. The qualitative results in Fig.~\ref{fig:pref_clip} show the same trade-off: semantics improve, while perceptual quality drops mildly. Importantly, unlike pure score maximization, this setting does not exhibit obvious reward hacking. This suggests that pairwise preference fitting acts as a stabilizing regularizer in joint optimization, making it possible to incorporate auxiliary score rewards without collapsing to reward-hacked solutions.

\subsection{Robustness to Noisy Preferences}
To test robustness to imperfect preference supervision, we inject synthetic noise into the pairwise reward during training by independently flipping each pairwise outcome with probability $p{=}0.1$, roughly matching typical human-disagreement rates on image preferences.

As shown in Tab.~\ref{tab:pref_grpo_noise}, \ourmethod remains strong under noisy preferences and consistently outperforms all point score-based baselines. Although the injected noise causes a small drop relative to the clean \ourmethod model, the degradation is limited across all metrics. This result suggests that pairwise preference optimization remains robust under moderate label corruption, which is important in practical settings where preference annotations are often noisy, inconsistent, or partially unreliable.

\begin{table}[t]
  \centering
  \scriptsize
  \setlength{\tabcolsep}{2pt}
  \renewcommand{\arraystretch}{1.08}
  \caption{\textbf{Robustness of \ourmethod under 10\% preference noise}. The noisy variant flips each pairwise preference outcome with probability $p{=}0.1$ during training.}
  \vspace{-6pt}
  \label{tab:pref_grpo_noise}
  \begin{tabular*}{\columnwidth}{@{\extracolsep{\fill}}lccccc@{}}
    \toprule
    \textbf{Method} & \textbf{UniGen} & \makecell{\textbf{T2I-}\\\textbf{Comp}} & \textbf{GenEval} & \makecell{\textbf{Unified}\\\textbf{Reward}} & \makecell{\textbf{Pick}\\\textbf{Score}} \\
    \midrule
    FLUX.1.dev & 61.30 & 48.17 & 62.92 & 3.04 & 22.42 \\
    w/ HPSv2+CLIP & 58.77 & 46.77 & 59.31 & 3.09 & 22.62 \\
    w/ UnifiedReward & 63.62 & 50.20 & 67.28 & 3.14 & 22.88 \\
    \midrule
    \makecell[l]{w/ Pref-GRPO\\(10\% noise)} & \underline{67.92} & \underline{51.03} & \underline{70.12} & \underline{3.22} & \underline{22.90} \\
    \rowcolor[HTML]{E2F4E3} w/ Pref-GRPO (\textbf{Ours}) & \textbf{69.46} & \textbf{51.85} & \textbf{70.53} & \textbf{3.26} & \textbf{23.02} \\
    \bottomrule
  \end{tabular*}
    \vspace{-0.2cm}
\end{table}

\subsection{Human Validation of the \ourbench Evaluator}
To verify the reliability of the automatic judge used in \ourbench, we conduct a testpoint-level agreement study between Gemini-2.5-Pro~\cite{huang2025gemini} and human annotators. We sample 400 held-out prompt-image-testpoint triplets from benchmark predictions generated by models of varying strengths, with 40 triplets for each of the 10 primary dimensions. Each triplet contains a prompt, a generated image, and one explicit testpoint description. Five annotators independently judge whether the image satisfies the specified testpoint using the same binary protocol as in benchmark evaluation, and the human majority vote is taken as the reference label.

We report inter-annotator agreement, Gemini-vs.-human-majority accuracy, macro-F1, and Cohen's $\kappa$ in Tab.~\ref{tab:gemini_human_validation}. These metrics provide complementary views of evaluator reliability: inter-annotator agreement reflects the intrinsic ambiguity of each dimension and serves as a practical human reference ceiling, Gemini-vs.-human-majority accuracy measures direct agreement with human judgment, macro-F1 prevents class imbalance from dominating the result, and $\kappa$ discounts agreement that may occur by chance. Gemini-2.5-Pro achieves strong overall consistency with human judgments. Agreement is highest on more direct dimensions such as \textit{Style}, \textit{Attribute}, and \textit{Layout}, while more abstract dimensions such as \textit{Relationship}, \textit{Compound}, and \textit{Logical Reasoning} remain relatively more challenging. This tracking behavior supports Gemini-2.5-Pro's use as the \ourbench evaluator.

\begin{table}[t]
\centering
\scriptsize
\setlength{\tabcolsep}{2.0pt}
\renewcommand{\arraystretch}{1.08}
\caption{\textbf{Human validation of Gemini-2.5-Pro as the \ourbench evaluator}. We report Inter-Ann. Agr., Gemini-vs.-human-majority Acc., Macro-F1, and Cohen's $\kappa$ for each primary dimension.}
\vspace{-6pt}
\begin{tabularx}{\columnwidth}{>{\raggedright\arraybackslash}p{1.95cm}YYYY}
\toprule
\textbf{Dimension} & \makecell{\textbf{Inter-Ann.}\\\textbf{Agr.}} & \makecell{\textbf{Acc.}} & \makecell{\textbf{Macro-}\\\textbf{F1}} & $\boldsymbol{\kappa}$ \\
\midrule
Overall & 0.92 & 0.83 & 0.82 & 0.66 \\
Style & 0.96 & 0.90 & 0.89 & 0.76 \\
World Know. & 0.91 & 0.82 & 0.81 & 0.64 \\
Attribute & 0.93 & 0.84 & 0.83 & 0.67 \\
Action & 0.91 & 0.82 & 0.81 & 0.64 \\
Relationship & 0.89 & 0.80 & 0.79 & 0.60 \\
Compound & 0.88 & 0.79 & 0.78 & 0.58 \\
Grammar & 0.86 & 0.81 & 0.77 & 0.60 \\
Layout & 0.92 & 0.83 & 0.82 & 0.66 \\
Logical Reason. & 0.85 & 0.79 & 0.77 & 0.60 \\
Text & 0.88 & 0.84 & 0.83 & 0.66 \\
\bottomrule
\end{tabularx}
\label{tab:gemini_human_validation}
\vspace{-0.2cm}
\end{table}

\section{Conclusion}

This paper addresses two limitations in current T2I research: unstable GRPO optimization under pointwise reward score maximization, and the lack of fine-grained benchmarks for diagnosing semantic consistency. To this end, we propose \ourmethod, which replaces absolute score maximization with pairwise preference fitting, and \ourbench, which evaluates semantic consistency through fine-grained testpoints spanning diverse prompt themes and evaluation dimensions.
Extensive experiments show that \ourmethod yields more stable training, alleviates reward hacking, and achieves consistently stronger semantic consistency on both in-domain and out-of-domain benchmarks. Besides, \ourbench reveals where strong T2I models still fail, especially on more demanding dimensions such as relationship understanding, compositional instruction following, logical reasoning, and text rendering. We hope these two contributions together provide a more reliable training objective and a more informative evaluation protocol for future T2I reinforcement learning research.

\bibliographystyle{IEEEtran}
\bibliography{example_paper}

\clearpage

\end{document}